\pdfoutput=1

\documentclass[a4paper,fleqn]{cas-dc}

\usepackage[T1]{fontenc}  
\usepackage[utf8]{inputenc} 
\usepackage{amsmath,amssymb,amsfonts} 
\usepackage[numbers]{natbib}
\usepackage{parskip}
\usepackage{pdfpages}
\usepackage{tabularray}
\usepackage{caption} 
\usepackage{graphicx}
\usepackage{longtable} 
\usepackage{array}
\usepackage{multicol} 
\usepackage{booktabs}
\usepackage{color}
\usepackage{pdflscape}
\usepackage{makecell}
\usepackage{float}
\usepackage{ragged2e} 
\usepackage{rotating}
\usepackage{adjustbox}
\usepackage{needspace}
\renewcommand{\arraystretch}{1.2}
\definecolor{headerpink}{rgb}{1,0.85,0.95} 

\def\tsc#1{\csdef{#1}{\textsc{\lowercase{#1}}\xspace}}
\tsc{WGM}
\tsc{QE}
\tsc{EP}
\tsc{PMS}
\tsc{BEC}
\tsc{DE}

\begin{document}
\let\WriteBookmarks\relax
\def\floatpagepagefraction{1}
\def\textpagefraction{.001}
\shorttitle{Reinforcement Learning as the Defining Leap in Healthcare AI}
\shortauthors{D. Perera et~al.}

\title [mode = title]{Beyond Prediction: \\Reinforcement Learning as the Defining Leap in Healthcare AI}



\tnotetext[1]{This research
  was supported by the the Talent Development Award 2023 from Saw Swee Hock School of Public Health, grant number 24-0180-A0001-0.}


\author[1]{Dilruk Perera}
\ead{dilruk@nus.edu.sg}
        
\author[1]{Gousia Habib}
\cormark[1]
\ead{gousiya1@nus.edu.sg}

\author[1]{Qianyi Xu}[style=chinese]
\ead{wjh@example.org}

\author[1]{Daniel J. Tan}
\ead{djtan@nus.edu.sg}

\author[1]{Kai He}
\ead{kai_he@nus.edu.sg}

\author[2]{Erik Cambria}
\ead{cambria@ntu.edu.sg}

\author[1]{Mengling Feng}
\ead{ephfm@nus.edu.sg}

\affiliation[1]{organization={National University of Singapore},
        addressline={21 Lower Kent Ridge Rd}, 
        postcode={119077}, 
        country={Singapore}}

\affiliation[2]{organization={Nanyang Technological University},
        addressline={50 Nanyang Ave}, 
        postcode={639798}, 
        country={Singapore}}

\cortext[cor1]{Corresponding author}

{\color{red}
\begin{abstract}
Reinforcement learning (RL) marks a fundamental shift in how artificial intelligence is applied in healthcare. Instead of merely predicting outcomes, RL actively decides interventions with long term goals. Unlike traditional models that operate on fixed associations, RL systems learn through trial, feedback, and long-term reward optimization, introducing transformative possibilities and new risks. From an information fusion lens, healthcare RL typically integrates multi-source signals such as vitals, labs clinical notes, imaging and device telemetry using temporal and decision-level mechanisms. These systems can operate within centralized, federated, or edge architectures to meet real-time clinical constraints, and naturally span data, features and decision fusion levels. This survey explore RL's rise in healthcare as more than a set of tools, rather a shift toward agentive intelligence in clinical environments. We first structure the landscape of RL techniques including model-based and model-free methods, offline and batch-constrained approaches, and emerging strategies for reward specification and uncertainty calibration through the lens of healthcare constraints. We then comprehensively analyze RL applications spanning critical care, chronic disease, mental health, diagnostics, and robotic assistance, identifying their trends, gaps, and translational bottlenecks. In contrast to prior reviews, we critically analyze RL’s ethical, deployment, and reward design challenges, and synthesize lessons for safe, human-aligned policy learning. This paper serves as both a a technical roadmap and a critical reflection of RL’s emerging transformative role in healthcare AI not as prediction machinery, but as agentive clinical intelligence.
\end{abstract}
}

\maketitle
\section{Introduction}
\begin{small}
\begin{table}[H]
\centering
\scalebox{0.9}{
\begin{tabular}{c l }
\toprule
\textbf{Abbreviations} & \textbf{Full Form } \\ \hline
RL           & Reinforcement Learning \\ 
MBRL          & Model Based RL \\
MFRL          & Model Free RL \\
DTR           & Dynamic Treatment Regimes \\
EHR          & Electronic Health Records \\ 
MDP           & Markov Decision Process \\
POMDP         & Partially Observable Markov Decision Process \\ 
DRL           & Deep RL \\
DQN          & Deep Q-Network \\ 
DDPG          & Deep Deterministic Policy Gradient \\
SAC          & Soft Actor-Critic \\ A3C           & Asynchronous Advantage Actor-Critic \\
MARL          & Multi-Agent RL \\ GEN-RL         & Generative RL \\
FED-RL         & Federated RL \\ GHE-RL         & Global Health Equity RL \\
BIRL          & Bayesian Inverse RL \\ CQL           & Conservative Q-Learning \\
BCQ          & Batch-Constrained Q-Learning \\ LSTM          & Long Short-Term Memory \\
PPO           & Proximal Policy Optimization \\
GAN          & Generative Adversarial Network \\ Q-Learning       & A RL algorithm \\
HIRL          & Hierarchical Reinforcement Learning\\
LR           & Logistic Regression\\
KUH          & Kupio University Hospital\\
WTRL          & Wavelet Transformed Reinforcement Learning\\
ES-RL         & Early Stop Reinforcement Learning\\
SER          & Safe Expeeriece Reshaping\\
RF           & Random Forest\\
\bottomrule
\end{tabular}
}
\end{table} 
\end{small}

Reinforcement Learning (RL) is poised to redefine how artificial intelligence interfaces with clinical care. While traditional machine learning focuses on retrospective predictions, RL introduces a fundamentally different ambition: learning to make decisions that optimize patient outcomes over time. Artificial intelligence (AI) has become an integral part of across diverse sectors including finance, transportation, education, logistics, and scientific research, through rapid advances in data availability, computing power and machine learning algorithms. In healthcare, AI is frequently cited as a catalyst for transformation, with national policy strategies, industrial investments and academic road maps emphasizing its potential to reshape clinical workflows and improve health outcomes~\cite{wu2025harnessing,lin2025has,he2025survey}. To date, the primary advances have been achieved from supervised learning, particularly in tasks such as disease diagnosis~\cite{kumar2023artificial, wu2023megacare,he2019understanding}, risk stratification~\cite{ALBAHRI2023156}, and early warning systems~\cite{macintyre2023artificial}. These models have demonstrated superior clinical utility in areas including diagnostic imaging (e.g., diabetic retinopathy)~\cite{panayides2020ai}, outcome forecasting (e.g. hospital readmissions)~\cite{mohanty2022machine,wu2024promise}, and clinical documentation support~\cite{bongurala2024transforming}. However, these successes, while impactful, showcase incremental improvements to existing paradigms instead of fundamental transformations in the delivery or organization of care~\cite{gu2023beyond}.

Despite the considerable advances of supervised learning based healthcare prediction tasks, they remain inherently limited in its ability to deliver the transformative potential promised by AI in healthcare. Supervised learning depend on static datasets and predefined labels, optimizing for a single-step predictions such as classification or regression~\cite{an2023comprehensive}. In contrast, healthcare delivery is inherently dynamic and temporally extended~\cite{xie2022deep}, involving sequences of interdependent decisions taken over time, often under uncertainty and with delayed outcomes~\cite{yu2021reinforcement}.

These temporal and causal dependencies are poorly captured by models trained on fixed snapshots of data~\cite{gao2023causal}. For example, static treatment protocols often fail to adapt to individual patient responses, and conventional dosing regimens are unable to adopt for real-time physiological fluctuations. From titrating medications in the ICU and scheduling patient appointments to optimizing resource allocation and coordinating care transitions, each decision has downstream effects that cannot be effectively modeled using single-step predictions~\cite{gommers2008medications}. Healthcare tasks span treatment optimization, care coordination, resource management, workflow scheduling, and patient engagement, requiring adaptive strategies that evolve based on changing circumstances and feedback. Moreover, conventional systems often require accurate ground truth data, which is difficult to obtain due to incomplete information, disorganized formats, and privacy restrictions, further limiting their applicability in real-time healthcare scenarios~\cite{tayefi2021challenges}. As a result, conventional supervised learning based AI systems that excel at answering ``what will happen''? remain fundamentally incapable of addressing the more critical question: ``what should be done?''~\cite{Shickel2018DeepEHR}.

RL~\cite{sutton1998introduction} represents the next step in the healthcare AI evolution, the truly transformative paradigm capable of delivering on AI's revolutionary promise in healthcare. Unlike the predictive limitations of supervised learning, RL by design learns how to act and optimize across diverse healthcare processes~\cite{yu2021reinforcement}, fundamentally shifting from passive prediction to active decision-making. RL addresses the core limitations of existing approaches by enabling adaptive decision-making through trial-and-error learning, where agents interact with their environment to optimize long-term outcomes while accounting for delayed effects of actions~\cite{Shickel2018DeepEHR,jayaraman2024primer}. Unlike supervised approaches, RL is designed to learn decision policies through continuous interactions with the environment, guided by long-term objectives. This paradigm directly addresses the fundamental limitations of snapshot-based predictions by learning how current actions influence future states over extended time horizons~\cite{shortreed2011informing}. By design, RL explicitly incorporates temporal dynamics and sequential decision-making, making it uniquely suited for healthcare settings where interventions, workflows, and resource allocation decisions that are continuous and must account for both immediate effects and long-term consequences~\cite{sutton1999reinforcement}. Furthermore, RL models dynamically learn from experience and delayed rewards without the need for labeled outcomes for every decision point, naturally aligning with the iterative and adaptive healthcare delivery~\cite{komorowski2018artificial}.

RL has demonstrated remarkable efficacy in high dimensional and sequential decision-making domains, including strategic gameplay~\cite{silver2016mastering}, robotic manipulation~\cite{levine2016end}, and autonomous navigation~\cite{kendall2019learning}. These successes showcase RL's capability in environments characterized by uncertainty, feedback-driven learning, and complex temporal dependencies. Given the similar characteristics in the healthcare setting, RL is increasingly being explored as a technique for developing adaptive policies across the healthcare spectrum, from personalized treatment protocols and care pathway optimization to resource allocation, staff scheduling, and patient flow management~\cite{gottesman2019guidelines,yu2021reinforcement}. RL-based models can dynamically adjust chemotherapy dosing to reduce tumor size over time~\cite{yang2023reinforcement}, optimize ventilation strategies in ICUs to improve survival rates~\cite{prasad2017reinforcement}, enhance anomaly detection in medical imaging~\cite{zha2020meta}, and streamline hospital operations such as bed allocation and emergency department triage~\cite{wattanapanit2025reinforcement}.

When applied to healthcare, RL promises a foundational shift the use of AI in healthcare from passive risk estimation to active decision optimization, allowing perspective analytics that aim to improve outcomes by shaping decisions in real time~\cite{gottesman2019guidelines}. Unlike supervised learning models that focus on static outcome prediction, RL explicitly models how each decision (e.g., clinical interventions, resource allocation decisions, workflow optimizations, or care coordination strategies) affects future system states and long-term outcomes~\cite{sutton1998introduction,komorowski2018artificial}. This capability allows RL to learn policy-level strategies that adapt to evolving healthcare contexts, capturing temporal dependencies and cumulative effects across patient care and operational efficiency~\cite{nemati2016optimal}. However, this transition from prediction to action introduces significant new challenges and risks. RL models may inherit biases from observational data, optimize for reward functions that are misaligned with healthcare goals, or produce policies that lack interpretability~\cite{adlung2021machine}. Moreover, when RL models appear to outperform existing practices in retrospective simulations, it raises critical questions about whether it reflects improved decision-making or complex imitation of potentially flawed historical data~\cite{zhu2023offline}.

Early applications of RL in healthcare have demonstrated both the potential and challenges of this approach across diverse settings, including clinical decision-making for sepsis treatment~\cite{komorowski2018artificial}, glucose control in type 1 diabetes~\cite{perera2025smart}, mechanical ventilation strategies~\cite{kondrup2023towards}, as well as operational applications such as emergency department patient flow optimization~\cite{wu2025reinforcement} and ICU resource allocation~\cite{ali2022reinforcement}. Initial efforts relied heavily on tabular methods, strong markovian assumptions, and small, simplified datasets, limiting generalizability and interpretability. Recent advances in offline RL algorithms enable learning safer and scalable policies from large retrospective datasets, avoiding the ethical and logistical challenges of real-time experimentation~\cite{gottesman2019guidelines}. Algorithms such as batch-constrained deep Q-learning (BCQ)~\cite{fujimoto2019off}, conservative Q-learning (CQL)~\cite{kumar2020conservative}, and importance-weighted evaluation techniques~\cite{jiang2016doubly,farajtabar2018more} have significantly improved the stability, policy reliability and practicality of applying RL across healthcare domains. In parallel, advancements in model-based RL~\cite{yu2020mopo}, healthcare-specific policy regularization~\cite{gottesman2019guidelines}, and generalization across patient subgroups have further improved sample efficiency, reduced overfitting, and improved interpretability across both clinical and operational healthcare settings.

Despite the algorithmic progress in RL, realizing its full potential in healthcare requires systmatic alignment with clinical objectives, operational constraints, and ethical standards across the healthcare ecosystem~\cite{wiens2019no}. Key technical challenges include designing clinically meaningful and operationally feasible reward structures~\cite{riachi2021challenges}, ensuring generalizability of learned policies across heterogeneous health systems and use cases~\cite{ghassemi2021false}, achieving secure, real-time integration into electronic health record and hospital information systems~\cite{rajkomar2018scalable}, and establishing medico-legal accountability for autonomous decision support across clinical and operational domains~\cite{amann2020explainability}. Additionally, privacy and regulatory constrains complicate data availability and exchange, requiring rigorous safeguards to address bias, fairness and representativeness~\cite{char2018implementing}. The opaque, black-box nature of many RL models also poses a considerable barrier to clinical deployment, necessitating the need for interpretable and transparent learning systems to gain the trust of healthcare providers and patients~\cite{tonekaboni2019clinicians}.

This survey provides a comprehensive and critical review of RL as a paradigm shift in healthcare AI, moving beyond passive risk prediction towards active, adaptive decision making. By learning policies that directly optimize for long-term clinical or operational goals, RL provides a prescriptive alternative to traditional supervised learning approaches, positioning RL as a foundational framework for building intelligent, autonomous healthcare systems.

We begin by outlining foundational RL concepts in Section 3, covering model-free, model-based, and offline approaches, and emerging emerging directions such as Recurrent RL, Hierarchical RL, and Inverse RL, discussing their theoretical and practical adaptations to healthcare environments. Section 4 systematically explores RL's applications across the healthcare domain, from personalized treatment planning, surgical optimization, and critical care to diagnostic support and robotic interventions. Section 5 reviews key challenges, spanning from data limitations, reward formulation, fairness, transparency, and regulatory accountability, need to be considered to safely deploy RL in real-world healthcare settings. Section 6 discusses the trade-offs underlying RL model design, including sample efficiency, reward alignment, and off-policy evaluation in partially observable clinical contexts. The Section 7 highlights emerging trends, such as multi-agent RL, federated RL, generative modeling, and human-in-the-loop systems, each offering new opportunities for scalable, collaborative and ethically grounded RL applications. Finally, Section 8 concludes with key insights and recommendations for advancing trustworthy, effective, and scalable RL deployment throughout the healthcare ecosystem.

To the best of our knowledge, this is the most comprehensive survey since 2020 dedicated to RL in healthcare. It integrates theoretical foundations with real-world clinical and operational implications, providing a critical taxonomy, synthesized evaluations, and a forward-looking agenda. Our aim is to support the safe, effective, and responsible use of RL as a transformative paradigm for intelligent, autonomous decision making in life-critical healthcare systems.

This survey is designed to serve a multidisciplinary audience, from AI researchers seeking to apply RL in healthcare, clinicians exploring decision-support technologies, and healthcare administrators and policymakers evaluating the feasibility of autonomous systems. Our aim is to provide technical foundation to RL, compile the state of RL in healthcare, critically evaluate its practical and ethical limitations, and provide a forward-looking roadmap supporting methodological innovation with real-world clinical and operational demands.The overall Taxonomy of the survey paper is depicted in Figure 1.

Throughout this survey, we map RL components into different fusion levels (i.e., data, feature or decision) and to fusion architecture (sensor layer, preprocessing fusion/policy, decision/activation) highlighting how fusion choices affect safety, sample-efficiency and transportability.

\begin{figure*}
  \centering
  \includegraphics[scale=0.53]{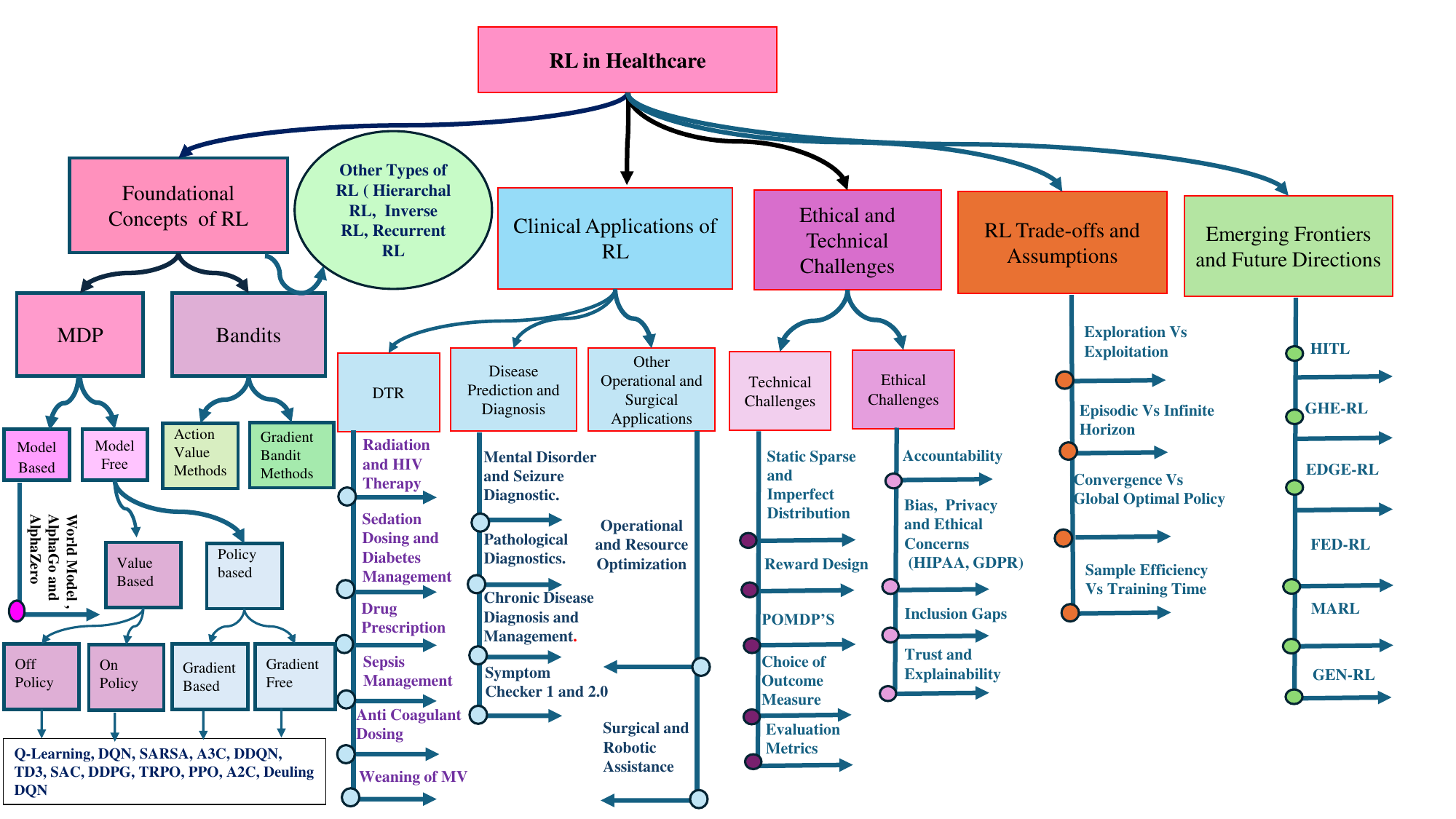}
  \caption{\textbf{Taxonomy of Reinforcement Learning in Healthcare.} The figure illustrates the core branches covered in this survey, including RL foundations, clinical and operational applications, technical and ethical challenges, and emerging frontiers such as generative RL, federated learning, and multi-agent systems. This taxonomy reflects RL’s evolution from prediction to action and provides a conceptual roadmap for the rest of the survey.}
  \label{F13}
\end{figure*}
\section{Methodology (Search Strategy)}
This section outlines the systematic search and selection technique used to identify high-quality studies relevant for RL in healthcare. The search method was structured to curate a representative and rigorous body of literature that covers both methodological innovation and real-world applicability across clinical and operational healthcare domains.

\subsection{Search Strategy}
A structured literature review was conducted across various databases: PubMed, BioMed IEEE Xplore, Google Scholar, MEDLINE, ScienceDirect, ACM Digital Library, SpringerLink, Elsevier, The Lancet, Nature Medicine and Web of Science, covering publications from year 2020 to 2025. The search combined general and domain-specific terms, including ``RL in healthcare'', ``deep RL in healthcare'', ``RL in ICU'', ``Sepsis Treatment'', ``RL for MV'', ``RL for drug dosing'', and related phrases. The full search string combinations are detailed in Appendix A Figure~\ref{F9}.
\subsection{Inclusion and Exclusion Criteria}
We included papers if they (1) focused on the application, evaluation, or development of RL methods in healthcare, (2) used patient data or simulated clinical environments, and (3) were published in English between 2020 and 2025.

Moreover, we made exceptions to include 16 additional papers (mentioned in Section 6) that falls outside the strict inclusion criteria as they offer critical insights into challenges such as in reward specification, exploration-exploration trade-offs, and off-policy evaluation. These works provide valuable conceptual insights into RL robustness, safety and ethical considerations in clinical application.
\subsection{Data Extraction and Synthesis}
From the final corpus, data were manually extracted based on key characteristics such as study purpose and scope, data source and sample size, RL algorithms used, evaluation frameworks, key findings and limitations. Due to the lack of established reporting standards for RL articles, no formal quality assessment was conducted. Instead, studies were thematically grouped by clinical task (e.g., treatment optimization, diagnosis) and technical innovation (e.g., offline RL, model-based RL) for subsequent analyzes.
\subsection{Selection Process and Results}
We initially identified a total of 350 full-text articles. After independent screening and consensus review by all co-authors, 246 articles were selected for the final review. 104 articles were discarded due to irrelevance to the survey paper. The selection process was illustrated using the PRISMA flow diagram in Appendix A Figure~\ref{F9}. 
\section{RL Theoretical Background and Foundational Concepts} 

RL framework allows agents to learn optimal behaviours through sequential interaction with dynamic environments. This learning paradigm shifts from standard static and supervised prediction models to optimize actions over time, aligning more closely with the complex, longitudinal nature of healthcare decision-making. This section provides a concise yet comprehensive overview of the theoretical foundations of RL, with a special focus on healthcare applications. Readers familiar with RL fundamentals may skip this section.
\subsection{Overview of RL}
RL is a subset of machine learning where an intelligent agent learns optimal actions through interaction with a dynamic environment. Unlike supervised learning, RL does not require labeled datasets; instead, the agent learns by continuously receiving feedback in the form of rewards from its environment. RL suits healthcare scenarios with sequential decision-making, uncertainty, and long-term goals. In healthcare, this shift enables prescriptive analytics (i.e., models that not only assess patient risk but also recommend actions based on anticipated long-term outcomes) transforming how treatment strategies, care pathways, and operational decisions are derived.

Mathematically, this sequential decision-making process is called the Markov Decision Process (MDP)~\cite{howard1960dynamic}. The Markov Property states that the future state of a process depends only on the current state and not on the sequence of events that preceded it. An RL task that satisfies the Markov property is called an MDP, which is defined by state and action pairs. Given a state $s$ and action $a$, the probability of each possible pair of the next state and reward, $s'$, $r$, is denoted as
\begin{small}
  \begin{equation}
  p(s', r \mid s, a) = \Pr\{S_{t+1} = s', R_{t+1} = r \mid S_t = s, A_t = a\}.
\end{equation}
\end{small}
Given the dynamics, we can compute the reward as,
\begin{small}
  \begin{equation}
r(s, a) = \mathbb{E}[R_{t+1} \mid S_t = s, A_t = a] = \sum_{r \in \mathbb{R}} r \sum_{s' \in \mathbb{S}} p(s', r \mid s, a)
\end{equation}
\end{small}
and state-transition probability as,
\begin{small}
  \begin{equation}
  p(s' \mid s, a) = \Pr\{S_{t+1} = s' \mid S_t = s, \, A_t = a\} = \sum_{r \in \mathbb{R}} p(s', r \mid s, a)
\end{equation}
\end{small}
This is the probability of the next state, given the current state and action, which represents an environment for the agent to interact with.

In the clinical setting, RL offers great potential in optimizing healthcare problems. The RL components of a typical MDP and the corresponding counterparts in clinical settings are:
\begin{itemize}
  \item \textbf{State} $S$: At each time $t$, the agent receives a representation of the environment $s_t$. To optimize the dynamic treatment regime, the state represents the patient's condition, which involves patient demographics, vital signs, lab tests, and other biomarkers~\cite{khezeli2023reinforcement, lu2021deep}. representative features commonly used in RL-based patient state representations are summarized in Table~\ref{tab:features}. For disease treatment, states may include patients' symptoms and any previous diagnostic steps taken~\cite{gottesman2019guidelines}.
  \item \textbf{Action} $A$: At each time $t$, the agent takes an action $a$. In disease treatment recommendations, it can be any intervention given by the clinicians, depending on the type of disease, like sepsis treatment (IV fluids and vasopressors) and ventilation settings (Fio2, PEEP, and tidal volume)~\cite{liu2020reinforcement}. For a surgical robot, actions might be specific movements~\cite{haiderbhai2024sim2real}.
  \item \textbf{Reward} $r$: The reward signal thus defines the good and bad events for the agent. Usually, there are both short-term and long-term rewards to compensate for the property of delayed feedback in healthcare. The short-term rewards might be based on acuity scores, reflecting immediate health improvements, while long-term rewards focus on terminal outcomes like survival or recovery duration.
  \item \textbf{Policy} $\pi$: A policy defines the learning agent’s behavior at a given time. Roughly speaking, a policy is a mapping from perceived states of the environment to actions to be taken in those states, like a model suggesting the best treatment.
\end{itemize}
\begin{small}
  \begin{table}
  \centering
  \caption{Examples of Clinical Features for Patient States}
  \label{tab:features}
  \begin{tabular}{|m{2cm} | m{5.5cm}|}
    \toprule
    \textbf{Category} & \textbf{Features} \\
    \midrule
    \textbf{Demographics} & Age, Race, Gender, Weight, Height, Re-admission Status \\ \hline 
    \textbf{Vital Signs} & Heart Rate, Mean BP, Systolic BP, Diastolic BP, Respiratory Rate, Body Temperature, SpO2, Shock Index, GCS (Glasgow Coma Scale) \\ \hline 
    \textbf{Laboratory Tests} & Arterial pH, Bicarbonate (HCO3), Base Excess, Glucose, Lactate, BUN,  Creatinine, Bilirubin, Magnesium, Sodium, Potassium, Chloride, Ionized Calcium, Calcium, Haemoglobin, WBC Count, Platelet Count, INR, PT (Prothrombin Time), PTT (Partial Thromboplastin Time), SGOT, SGPT, Albumin, CO2, ALAT, Anion Gap, ASAT, Hematocrit, Prothrombin Time, FiO2, PaO2, PaCO2, PaO2/FiO2 Ratio \\ \hline 
    \textbf{Ventilation} & Mechanical Ventilation, Arterial Blood Gas, Sedation \\ \hline 
    \textbf{Fluid Balance} & Fluid Input (4-hour/total), Fluid Output (4-hour/total), \\ \hline 
    & Cumulative Fluid Balance \\ \hline 
    \textbf{Clinical Scores} & SIRS, SOFA Score, Elixhauser, lODS \\ \hline 
    \textbf{Renal Support} & Renal Replacement Therapy \\ 
    \bottomrule
  \end{tabular}
\end{table}
\end{small}

From a fusion perspective, the ‘state’ is an information fusion result, combining synchronous/asynchronous sensor streams (e.g., telemetries), discrete lab events, imaging/text embeddings, and contextual priors. Policies and critics then operate on this fused representation. Conversely, decision-level fusion can also combine outputs from multiple policies (e.g., rule-based + RL, or clinician policy + learned policy) to improve safety and trust.

\subsection{MDP and POMDP} In a classic or fully observable MDP, the agent has complete knowledge of the environment's state at each time step. Most RL applications in healthcare assume an MDP framework, which is characterized as a tuple $(S, A, P, R, \gamma)$, where $S$ denotes the set of states (e.g., patient conditions), $A$ represents the set of actions (e.g., clinical interventions), $P(s'|s,a)$ signifies the transition probability between states, $R(s,a)$ indicates the reward function reflecting patient outcomes, and $\gamma$ is the discount factor for future rewards~\cite{al2024optimized}. In clinical settings, the observation at each step typically aggregates heterogeneous data sources (e.g., monitors, labs, notes, images) into a working state representation. The Markov Property assumes that future states rely solely on the present state and action, disregarding historical data. This simplification aids in modeling but may inadequately represent the intricacies of healthcare situations where patient information is frequently partially observable. Notwithstanding this, the MDP framework is extensively utilized in RL-based healthcare research owing to its efficacy in modeling sequential decision-making\cite {killian2020empirical}.

Clinical data are typically irregularly sampled and subject to uncertainty. The true state of a patient (e.g., disease progression, response to treatment) is often not directly observable. Instead, available features contains multi-modal, indirect observations of the latent patient state. Missing data, asynchrony across modalities and an incomplete understanding of biological and physiological processes, contribute to the partial observability of healthcare environments. Consequently, a growing body of RL research explicitly accounts for partial observability by incorporating more advanced state representations. Many methods leverage sequential models, such as recurrent neural networks (RNNs) and transformers, or employ ordinary differential equations (ODEs) to model state transitions. This has led to formulating healthcare decision-making problems as POMDPs~\cite{tsoukalas2015data, li2018actor, peng2018improving}. 

A POMDP is defined similarly to an MDP, with states, actions, and a transition function. Observation space can be defined as \( O = \prod_{m=1}^M O^{(m)}\) with modality index $m$ (e.g., vitals, labs, notes), which represents the set of possible observations the agent can receive instead of directly accessing the true states of the environment. The observation function \( Z(o | s', a) \) defines the probability of observing \( o \) given that the agent took action \( a \) and transitioned to state \( s' \), and, when appropriate, factored as $\prod_m Z^{(m)} (o^{(m)}|s',a)$ under conditional independence assumptions. 

Under partial observability, observations are inherently multi-modal: \(o_t=\{o_t^{(m)}\}_{m=1}^{M}\). Feature-level fusion encoders \(x_t=f_{\phi}(o_t,r_t)\) handle missingness and asynchrony, while recurrent or continuous-time models perform temporal fusion of \(x_t\) into beliefs \(b_t\). At the policy stage, decision-level fusion (e.g., ensembling or human-in-the-loop constraints) shapes \(a_t \sim \pi_{\theta}(b_t)\). These design choices (early/late/hybrid; centralized/federated/edge) leave the POMDP formalism intact but impact robustness and real-time feasibility.

\subsection{Types of RL Approaches}
This section discusses diverse RL methodologies, encompassing traditional approaches and sophisticated techniques such as deep RL (DRL), which employ deep neural networks for intricate decision-making problems.

\noindent\textbf{Deep RL:} RL is also known as DRL if it adopts deep neural networks (DNN) to replace tabular representations for $Q(s_t, a_t)$, $V(s_t)$ or $\pi$. Tabular forms are not feasible in complex environments whose states and actions could be infinite and not possible to be represented or implemented. In contrast, DNN could represent the relations between states and actions explicitly with nodes and weights. The network of nodes rearranges its weights based on the learning objective while evaluating training data through back-propagation.
DNNs are applied in RL in various ways. For instance, in policy gradient RL~\cite{alghanem2018asynchronous}, DNN can be used to optimize a policy $\pi(\theta)$, where $\theta$ is the parameters (weights) in DNN. In value-based RL, DNN can be used to approximate $V_{\phi}(s_t)$ or the Q-function $Q_{\phi}(s_t, a_t)$. In model-based RL~\cite{panayides2020ai}, DNN is used to approximate the transition function $p_{\phi}(s_{(t+1)} | s_t, a_t)$, where $\phi$ is the parameter in DNN that needs to be tuned through training.

\noindent\textbf{Policy-based RL:} A policy-based RL, such as a Policy Gradient RL, rolls out a random policy $\pi(\theta)$ in an environment and generates trajectories under the policy. After a few trials, the expected long-term reward $J(\theta)$ will be approximated by samples from the collected trajectories:
\begin{equation}
J(\theta) \approx \frac{1}{N} \sum_{i} \sum_{t} r(s_{i,t}, a_{i,t})
\end{equation}
Policy Gradient RL updates the policy $\pi_{\theta}$ by applying gradients to $J(\theta)$: $\theta' = \theta + \alpha \nabla_{\theta} J(\theta)$. This on-policy method requires new data after each policy update. A DNN constructs the policy, taking the state as input and outputting an action. The policy is learned by updating the DNN weights based on the gradient of $J(\theta)$. Policy Gradient RL is a type of on-policy[15] RL algorithm because every time the policy changes, we need to sample new data under this new policy. In contrast, off-policy learning learns from the data generated by a different policy. In off-policy RL, we can improve the current policy without generating new samples. This is convenient for many clinical scenarios when generating new data is expensive.

\noindent\textbf{Value-based RL:} 
Value-based RL estimates the value function \( V(s) \) or Q-function \( Q(s, a) \) using the Bellman equation. The objective is to minimize the MSE:
\[
L(\phi) = \frac{1}{2} \| \hat{V}_{\phi}^{\pi}(s_t) - y_{(t+1)} \|^2 \tag{5}
\]
For FVI and FQI, the target values are:
\begin{align*}
y_{(t+1)} &= \max_{a} \gamma Q^{\pi}(s_{(t+1)}, a),\\ 
\quad y_{(t+1)} & = r(s_t, a_t) + \gamma \max_{a} Q^{\pi}(s_{(t+1)}, a) \tag{6}
\end{align*}
In Q-learning, one tuple is used for approximation. Value-based RL is off-policy with low variance but faces challenges: correlated data, instability, and Q-function overestimation.

To address data correlation, Q-learning uses a replay buffer. Model instability is mitigated with a target network in Deep Q-Networks (DQN), where the target value is fixed for several iterations. Overestimation is addressed with Double Q-learning (Double DQN), using two DQNs: one for action selection and the other for evaluation.

\noindent\textbf{Actor-Critic RL:} Actor-Critic RL~\cite{heess2015learning} combines policy-based and value-based RL. State-action pairs are sampled from the policy, with the policy represented by a DNN (actor). The value function \( V_{\hat{\phi}}^{\pi}(s_t) \) is estimated by another DNN (critic). The advantage \( A_{\hat{\pi}}(s_i, a_i) \) measures how much better an action is than the average action:
\[
A_{\hat{\pi}}(s_i, a_i) = r(s_i, a_i) + V_{\hat{\phi}}^{\pi}(s') - V_{\hat{\phi}}^{\pi}(s_i) \tag{7}
\]
Using this advantage, the actor updates the policy, while the critic is optimized via supervised regression:
\begin{equation}
L(\phi)=\frac{1}{2} \lVert V_{\hat{\phi}}^{\pi}(s_t) - y_t \rVert^2 \label{8}  
\end{equation}
where \( y_t \) is estimated using either Monte Carlo or bootstrap methods~\cite{mnih2016asynchronous},~\cite{munos2016safe}. Notwithstanding the promise of Actor-Critic RL in healthcare, numerous obstacles emerge. These encompass data scarcity and variability, challenges in optimizing for long-term patient outcomes, partial observability of patient states, the exploration-exploitation dilemma, and ethical and regulatory issues on transparency and safety in clinical decision-making.

\noindent\textbf{Model-based RL:} All the discussed RL algorithms are model-free, where the transition function $p(s_{(t+1)} \mid s_t, a_t)$ is unknown. The RL agent cannot predict the next state and bypasses the transition function by sampling from the environment directly.
In model-based RL~\cite{panayides2020ai}, the goal is to learn the transition function. Given the current state and action, the algorithm estimates the probability of all possible next states. While model-based RL allows generating new samples, estimating the transition function can be challenging, especially for clinical applications. Commonly used model Gaussian processes(GP), DNNs, and Gaussian mixture models(GMM). In healthcare, learned world models also serve as fusion hubs, integrating multi-source trajectories into calibrated counterfactual simulators that enable safe policy search and off-policy evaluation. 

\noindent\textbf{Hierarchical RL:} When the learning task is large with multiple sub-optimal policies, Hierarchical RL~\cite{kulkarni2016hierarchical} uses a two-level structure. The lower level is similar to general RL algorithms, selecting an action \( a_t \) given state \( s_t \). In comparison, the higher level, the 'meta-policy', determines which lower-level policy to apply over a trajectory.

Hierarchical RL can learn a globally optimal policy faster and transfer knowledge from past tasks or lower-level policies. In clinical applications, where the state-action space is vast due to complex human behaviors, it is a natural choice. However, its architecture is complex, and inappropriate transfer learning can lead to 'negative transfer'~\cite{pan2009survey}, where the high-level policy may not outperform lower-level policies.

\noindent\textbf{Recurrent RL:} A key limitation of Markov decision processes is the Markov property, which is rarely satisfied in real-world problems. In medical applications, a patient’s full clinical state is often not observable, leading to POMDP~\cite{kaelbling1996reinforcement}. A POMDP is represented by a 4-tuple (S, A, R, O), where $O$ are observations. Classic DQN is effective only if observations match the underlying state. Hausknecht et al.~\cite{hausknecht2015deep} extended DQN by replacing the first fully connected layer with Long Short Term Memory (LSTM)~\cite{hochreiter1997long} to estimate \( S \) from \( O \).Their algorithm showed that it could integrate information successfully through time and could replicate DQN’s performance on standard Atari games with a setting of POMDP for the game screen. This modification led to the Deep Recurrent Q-Network (DRQN), which successfully integrates information over time and replicates DQN's performance on Atari games under a POMDP setting. 

\noindent\textbf{Inverse RL:} In most RL algorithms, reward functions are hand-crafted, which can be vulnerable to misspecification. Inverse RL provides an alternative by learning the reward function directly from expert demonstrations~\cite{ng2000algorithms}. While imitation learning learns from experts, it may lead to sub-optimal policies~\cite{abbeel2004apprenticeship} due to experts' varying capabilities. Inverse RL aims to recover the true reward function from sub-optimal demonstrations. A deep neural network (DNN) learns the reward function, where the input is state-action pairs $(s, a)$ from a sub-optimal policy $\pi^\#$, and the output is the reward $r_{\Phi}(s, a)$. This reward is then used to plan for a better, possibly optimal policy $\pi^*$, aligning with the clinician's goals. From a fusion perspective, IRL fuses heterogeneous demonstrations (units, sites, shifts) into a shared reward prior, making explicit how institutional context influences learned clinical preferences. 

Recent developments in RL have facilitated substantial progress in healthcare applications, classified according to policy kinds, RL paradigms, and model types. These encompass techniques such as DQN, Actor Critic approaches, and Inverse RL (IRL), each targeting distinct facets of decision-making and optimization.

The classification emphasizes essential ideas, including treatment optimization and learning from expert demonstrations. Despite these developments demonstrating encouraging results, obstacles persist, including generalization issues, ethical dilemmas, data quality concerns, and computing complexity. 

This methodology provides significant insights into how RL might enhance healthcare outcomes while pinpointing areas for additional advancement, as represented by Appendix B Table~\ref{Table-Recent-Advances}, which illustrates how these RL concepts are used in clinical use cases and how different algorithmic choices impact different clinical outcomes, learning settings, and real-world deployment challenges.

\section{Clinical Applications of RL in Healthcare}
RL's prescriptive nature makes it uniquely suited for healthcare settings where sequential decisions are made often under uncertainty with delayed outcomes.

This section surveys RL applications that demonstrates its ability to actively shape clinical trajectories, instead of merely classifying risks across a wide range of RL applications in healthcare. These applications include disease prediction, dynamic treatment protocols, surgical support, and the management of critical conditions such as sepsis and mechanical ventilation (MV). Specifically, we highlight key studies and milestones demonstrating how RL is transforming clinical decision-making and improving patient outcomes across various medical domains. Recent research demonstrated that RL can be effectively utilized to support clinical decision-making and treatment recommendations across various medical applications~\cite{nemati2016optimal, prasad2017reinforcement, tseng2017deep}. These include selecting medications and fluids, adjusting dosages for acute and chronic conditions, and determining the length of time patients require MV.

\begin{figure*} 
 \centering
 \scalebox{0.9}{
 \includegraphics[scale=0.55]{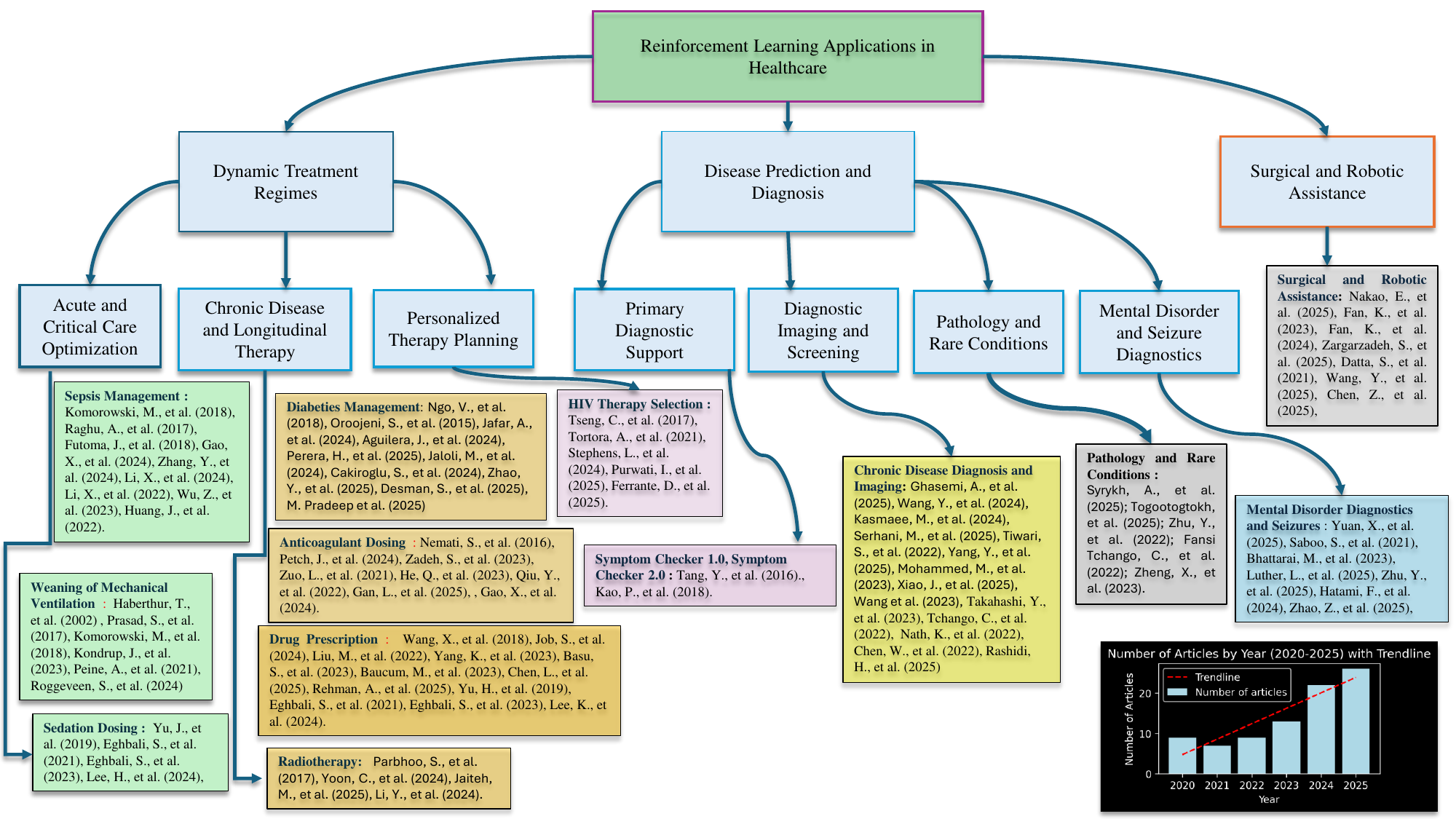} 
 }
 \caption{RL in Healthcare: A Mapping by Application Type.}
 \label{F1}
\end{figure*}

Figure~\ref{F1} represents an exhaustive review of the literature on RL applications in healthcare and mapping by application type. Classifying principal use cases like sepsis management, medicine dosing, surgical aid, chronic disease diagnosis, and others, alongside significant studies that have furthered progress in each domain. The preceding sections discuss the broader applications of RL in clinical settings, highlighting challenges and offering future insights.
\subsection{Dynamic Treatment Regimes}
Healthcare decision-making aims to create personalized treatment plans that adapt to a patient's changing health to improve long-term outcomes. Dynamic Treatment Regimes (DTRs) provide structured, evidence-based protocols that guide sequential clinical decisions such as therapy selection, dosing and timing based on patient conditions and history.\\
Unlike randomized controlled trials (RCTs), which validate treatment efficacy through controlled comparisons, they are typically static in nature, evaluating predefined treatment plans. In contrast, DTRS enable adaptive policies that optimize interventions over time for individuals and groups. RL is particularly well suited for learning such policies, as it allows modeling sequential, uncertain, and personalized decision-making. Specific application types can further organize these approaches in DTR.
\subsubsection{Acute and Critical Care optimization}
RL has demonstrated potential in enhancing critical care management, including sepsis therapy, mechanical ventilation weaning, and sedative dosing. RL models enhance personalized care through patient data, yet issues such as generalizability, interpretability, and clinical validation must be resolved to ensure real-world efficacy.

\noindent\textbf{Sepsis management:} Sepsis is a life-threatening condition triggered by an overwhelming immune response to infection, leading to systemic inflammation, organ dysfunction, and potentially death. Given the urgency and complexity of treatment, RL has emerged as a promising tool for personalized sepsis management. Komorowski et al. (2018) pioneered using model-based RL for optimizing intravenous fluid and vasopressor administration, leveraging retrospective ICU data to infer clinician-like treatment policies~\cite{komorowski2018artificial}. Building on this, Raghu et al. (2017) implemented a Double DQN architecture that reduced mortality by 1.8\%, although its reliance on discretized actions limited its clinical realism~\cite{raghu2017continuous}.

Futoma et al. (2018) addressed temporal dynamics by proposing a Deep Recurrent Q-Network (DRQN), achieving a potential 8.2\% reduction in mortality, though the model's interpretability remained limited~\cite{futoma2018learning}. Recent machine learning models have shifted focus towards prediction accuracy and interpretability. Gao et al. (2024) utilized random forests with SHAP analysis to predict sepsis mortality, achieving an AUROC of 0.94~\cite{gao2024prediction}. Similarly, Zhang et al. (2024)~\cite{zhang2024explainable}employed XGBoost with inflammatory biomarkers to develop a robust and interpretable model for in-hospital mortality prediction~\cite{zhang2024predicting}. In parallel, Li et al. (2022) proposed a collaborative Deep Q-Network to optimize glucose control in sepsis patients, simulating physician consultation through multi-agent learning~\cite{li2022electronic}.

To better align AI systems with clinical workflows, Wu et al. (2023) introduced WD3QNE, a Weighted Dueling Double Deep Q-Network that incorporates human expertise, achieving a survival rate of 97.81\%~\cite{Wu2023}. Complementing this, Huang et al. (2022) tackled the limitations of discrete dosing by implementing a continuous action RL model based on Deep Deterministic Policy Gradient (DDPG), improving treatment precision and adaptability~\cite{huang2022reinforcement}.

From a fusion perspective, it was evident that Sepsis policies typically fuse high-frequency monitors (MAP, SpO2), intermittent labs (lactate), and EHR context (comorbidities/med history) into a single state, and studies that add free-text embeddings or imaging tend to improve early recognition and dosing stability. Decision-level fusion (e.g., weighted clinician/RL policies) further stabilizes actions during hypotensive events.

Despite these advancements, challenges remain. Many models are not validated in real-time clinical settings and lack generalizability across diverse patient populations. Additionally, balancing interpretability with performance remains a key concern, underscoring the need for clinically grounded, adaptive RL frameworks that can operate reliably under real-world constraints.

\noindent\textbf{Weaning of MV:} Mechanical ventilation (MV) is a life-saving intervention used in approximately one-third of critically ill patients~\cite{haberthur2002extubation}. From a fusion perspective, essentially MV policies are multi-sensor problems where ventilator waveforms, gas exchange, sedation depth, and fluid balance must be fused despite different rates and lags. Studies that reward-shape with composite acuity scores implicitly perform objective fusion, aligning short-term gas targets with long-term outcomes. 

Current optimal weaning timing remains unclear, and inappropriate ventilation strategies can contribute to ventilator-induced lung injury. Recent studies have explored RL to personalize MV management by leveraging patient data and sequential decision-making frameworks. Prasad et al.~\cite{prasad2017reinforcement} developed an off-policy Fitted Q Iteration (FQI) algorithm for guiding weaning, achieving 85\% concordance with clinician decisions. Peine et al.~\cite{peine2021development} introduced Ventai, using tabular Q-learning over a large ICU dataset to optimize tidal volume, PEEP, and Fio2 settings. Their model demonstrated a 42.6\% improvement in estimated performance return over clinicians' policies and more frequent adoption of lung-protective strategies, including lower tidal volumes and optimized FiO2 levels.

Building on this, Kondrup et al.~\cite{kondrup2023towards} proposed DeepVent, a deep offline RL framework based on Conservative Q-Learning (CQL), which mitigated value overestimation and prioritized safety. They introduced intermediate rewards derived from APACHE II scores to address sparse terminal feedback, resulting in safer and more stable recommendations than Double DQN. Similarly, Roggeveen et al.~\cite{roggeveen2024reinforcement} applied RL to optimize ventilation in COVID-19 patients, introducing policy restriction (“king-knight” structure) and cross-off-policy evaluation to improve generalizability and safety. Their use of delta-Q metrics allowed clinical comparison between physician and model actions, enhancing interpretability.

While these studies highlight the promise of RL in optimizing MV strategies, key limitations persist. Challenges include sparse and delayed rewards, variability in patient response, and overestimating Q-values in offline settings. Techniques like CQL and reward shaping have helped address these, yet generalizability across diverse ICU populations remains limited. Future work must emphasize clinically validated reward functions, robust off-policy evaluation, and interpretable, safe policies to ensure real-world applicability.

\noindent\textbf{Sedation Dosing:} Sedation management in intensive care is complex, requiring personalized dosing strategies to maintain patient safety and avoid complications such as over-sedation or delirium. Sedation control fuses RASS targets, MAP/HR stability, ventilator parameters, and drug history. Several works achieve safety by decision-level fusion overlaying guardrails/veto rules on top of the learned policy to manage drug accumulation and titration inertia. RL has emerged as a promising approach to optimize sedation protocols. Yu et al.~\cite{yu2019inverse} applied inverse RL to analyze Propofol infusion patterns, revealing that clinicians implicitly prioritize physiological stability. Expanding on this, Eghbali et al.~\cite{eghbali2021patient} developed a deep Q-network (DQN) model for sedation management using Propofol and Fentanyl, improving sedation stability by 29\% and achieving optimal sedation in 99.5\% of ICU stays. In a follow-up, Eghbali et al.~\cite{eghbali2023reinforcement} introduced a continuous-action RL model based on DDPG, which outperformed clinician policies by 8\% for sedation control and 26\% in maintaining mean arterial pressure (MAP), underscoring the benefit of multi-objective optimization.

More recently, Lee et al.~\cite{lee2024reinforcement} proposed the AID (Artificial Intelligence for Delirium) model to optimize dexmedetomidine dosing for delirium prevention. Their RL agent, trained and validated across over 2,700 ICU admissions, demonstrated superior estimated performance returns over clinician policies in internal and external cohorts. The AID policy reduced average dexmedetomidine doses and achieved a significant reduction in delirium incidence by maintaining dosing decisions closer to the target sedation range.

Despite their success, these models face challenges such as patient variability, sparse reward signals, and limited prospective validation. While inverse and off-policy RL approaches improve safety and adaptability, real-world implementation requires improved interpretability and alignment with clinical workflows. Future work must address these limitations through hybrid models, interpretable policies, and robust off-policy evaluation frameworks to ensure reliable, personalized sedation in critical care. Appendix B Table~\ref{tab:Summary1-a} provides a summary of key studies that have applied RL for clinical applications, specifically in the context of acute and critical care optimisation.
\subsubsection{Chronic Disease and Longitudinal Therapy}
Managing chronic diseases requires individualised, long-lasting treatment approaches. Conventional approaches frequently fail because of patient diversity. RL has demonstrated potential in optimising treatment for disorders such as diabetes, anticoagulant dosage, HIV, and cancer, providing more personalised and effective treatments. 

\noindent\textbf{Diabetes Management:} Diabetes management requires continuous monitoring of blood glucose levels. Traditional insulin dosing can be challenging, leading to suboptimal control. RL offers a dynamic approach to adjusting insulin doses, personalizing treatment for more effective interventions. RL has proven effective in optimizing insulin dosing and diabetes management.

Ngo et al.~\cite{ngo2018reinforcement} developed a value-based RL system to minimize glucose fluctuations, but its simplicity limits its real-world applicability. Oroojeni et al.~\cite{oroojeni2015reinforcement} used Q-learning for personalized insulin dosing, but frequent adjustments were still required, indicating challenges in real-time application. Jafar et al.~\cite{jafar2024personalized} improved glucose control during high-fat meals, reducing hypoglycemia and increasing daily steps by 19\% ($p < 0.001$), demonstrating RL’s ability to adapt to dietary changes. Similarly, Perera et al.~\cite{perera2025smart} addressed imperfect clinical data, achieving a 12.2\% reduction in high HbA1c rates, proving RL's robustness under real-world conditions. Jaloli et al.~\cite{jaloli2024basal} achieved 92\% time in the target glucose range and reduced hypoglycemia ($p \leq 0.05$), highlighting RL’s effectiveness in minimizing glucose variability. Zhao et al.~\cite{zhao2025safe} further advanced RL with a closed-loop artificial pancreas system, achieving an 87.45\% time in range (TIR) and improving glucose stability. 

Wang et al.~\cite{wang2023optimized} demonstrated RL's effectiveness in Type 2 diabetes, reducing glucose from 11.1 mmol/L to 8.6 mmol/L, while Desman et al.~\cite{desman2025glucose} minimized insulin errors post-surgery, improving patient outcomes. Despite these advancements, RL faces challenges in data variability, real-world validation, and model complexity. While RL shows promise in personalizing insulin dosing and optimizing glucose control, further refinement is needed for broader clinical application.
\subsubsection{Anticoagulant Dosing} Anticoagulant dosing, particularly for heparin and warfarin, is crucial for preventing adverse events like bleeding and thrombosis. Traditional dosing methods often fail to account for individual patient variability, while RL offers a dynamic, personalized solution. 

For heparin, Nemati et al.~\cite{nemati2016optimal} optimized dosing by maintaining apt levels within therapeutic ranges using RL. Qiu et al.~\cite{qiu2022latent} introduced a batch-constrained RL model, achieving an 87.7\% therapeutic level maintenance, outperforming Nemati et al.'s approach. This difference highlights the advantage of Qiu et al.'s model in reducing dosing errors, although both methods rely on high-quality patient data. 

In warfarin dosing, Petch et al.~\cite{petch2024optimizing} utilized deep RL to optimize INR levels for atrial fibrillation (AF) patients, outperforming traditional methods. In contrast, Sadjad et al.~\cite{zadeh2023optimizing} applied a PK/PD model with RL, demonstrating superior results in managing warfarin therapy. While both approaches showed promise, Petch et al.'s method emphasized real-time clinical application, while Sadjad et al.’s model focused more on pharmacokinetic predictions. 

Gan et al.~\cite{gan2025ai} utilized CURATE.AI for warfarin dosing, reducing INR deviations and improving safety. However, the variability in real-world data remains a challenge for such models. Similarly, Zuo et al.~\cite{zuo2021improving}, and He et al.~\cite{he2023model} demonstrated RL’s effectiveness in improving treatment outcomes for AF patients and optimizing heparin dosing, but their models also face issues with data quality and real-time monitoring. 

Despite RL's remarkable progress in anticoagulant dosing, obstacles such as data variability, model complexity, and real-world validation persist. Also, RL presents considerable promise for individualised therapy; additional refinement and extensive validation are necessary for clinical use.
\subsubsection{Drug Prescription} 
RL has been explored across a wide range of prescription tasks-from real-time ICU medication delivery to long-term disease management. However, these efforts differ significantly in data modality, risk setting, and learning constraints. In critical care, Wang et al. employed supervised RL to reduce ICU mortality by 4.4\%, demonstrating the feasibility of sequential drug decision-making. Job et al. extended this to vasopressor titration in ARDS, where action delays were explicitly modeled~\cite{wang2018supervised},~\cite{job2024optimal}. Both studies utilized policy learning under constrained environments but differed in reward design-Wang prioritized terminal outcomes, while Job emphasized intermediate hemodynamic stabilization.

For long-term care, Liu et al. introduced PPORank~\cite{liu2022deep}, integrating clinical and molecular features to personalize recommendations. Similarly, Yang et al. adapted RL for Traditional Chinese Medicine (TCM)~\cite{yang2023optimization}, addressing combinatorial prescriptions. While these approaches benefit from richer contextual inputs, they often suffer from sparse reward signals and interpretability issues, which are critical for adoption in traditional and integrative medicine settings.

Several studies~\cite{basu2023challenges},~\cite{baucum2023optimizing} tackled delayed effects and safety constraints head-on. Basu et al. proposed a PAE-POMDP formulation to correct delayed treatment feedback, which is important in chronic settings where outcomes manifest over time. Chen et al. introduced Safe Integral RL~\cite {chen2025event}, incorporating safety thresholds directly into the optimization loop for immunotherapy dosing, aligning model behavior with clinical risk aversion.

In behavioral health, Baucum et al. optimized treatment for substance use disorders~\cite{baucum2023optimizing}. This domain poses unique challenges due to behavioral variability and noisy observational data, underscoring the need for robustness and longitudinal modeling.

Across these works, there is a tradeoff between clinical interpretability and model performance. Safe RL methods (e.g., Chen) offer practical safety guarantees but may underperform in less constrained settings. Conversely, models like PPORank and PAE-POMDP show theoretical flexibility but require complex inputs or assumptions. Moreover, deployment remains limited due to reliance on retrospective data, lack of prospective validation, and concerns around generalization.

\noindent\textbf{HIV Therapy Selection:} HIV-1 treatment requires lifelong antiretroviral therapy (ART), but repeated use of the same drug combinations often leads to resistance. RL and machine learning (ML) have been applied to optimize ART selection, addressing the sequential nature of therapy and enhancing treatment outcomes.

Parbhoo et al.~\cite{parbhoo2017combining} combined kernel regression with RL for HIV therapy selection, showing improvements over traditional approaches. However, its reliance on kernel regression limits real-world scalability. Yoon et al.~\cite{yoon2024optimal} applied Double Deep Q-Network (DDQN) with experience replay, optimizing treatment schedules and reducing resistance. This model is more adaptable than Parbhoo et al.'s, yet balancing exploration and exploitation remains challenging.

Jaiteh et al.~\cite{jaiteh2025application} shifted focus to ML for predicting HIV testing behavior, improving testing rates, and indirectly supporting therapy optimization. Li et al.~\cite{li2024predictive} used ML to predict HIV-related mortality, achieving a c-index of 0.83 and improving early death prediction, which informs treatment planning. While Jaiteh et al.'s work enhances testing rates, Li et al.'s model provides more direct insights into mortality, influencing long-term ART decisions. 

These studies highlight RL and ML's potential in HIV care, with Parbhoo et al.'s and Yoon et al.'s models focusing on treatment optimization. At the same time, the research of Jaiteh et al.~\cite{jaiteh2025application} and Li et al.~\cite{li2024predictive} emphasize complementary aspects like testing behavior and mortality prediction. Despite the promise of these approaches, challenges such as data heterogeneity, model complexity, and clinical validation remain. Appendix B Table~\ref{tab:Summary2} summarizes key studies exploring the application of RL in chronic disease management and longitudinal therapy.
\subsubsection{Personalized Therapy Planning}
Effective therapy planning for diverse medical disorders requires accurate treatment optimization to enhance efficacy while minimizing negative effects. Conventional approaches frequently prove inadequate because of patient diversity. RL provides a dynamic method for personalizing treatment protocols and improving clinical outcomes by adjusting to the specific demands of individual patients.

\noindent\textbf{Radiotherapy:} Radiation therapy for lung cancer requires precise dose optimization to balance tumour control and minimize damage to healthy tissue. RL has emerged as an innovative approach to personalizing treatment regimens and improving clinical outcomes. Several studies have employed RL to optimize lung cancer radiotherapy, each contributing to advancing treatment while facing challenges in real-world application.

Tseng et al.~\cite{tseng2017deep} used a deep Q-network (DQN) to optimize thoracic irradiation, balancing radiation-induced pneumonitis (RP) and local control (LC). Trained on data from 114 patients, their approach achieved results comparable to clinicians by tackling data scarcity with Generative Adversarial Networks (GANs). While this approach showed promise, it was limited by its reliance on simulated data, which may not fully represent the variability of real-world clinical cases.

Tortora et al.~\cite {tortora2021deep} applied deep RL to fractionated radiotherapy for non-small cell lung cancer (NSCLC), achieving improved tumour control while minimizing damage to healthy tissue. This method advances traditional techniques but faces the challenge of generalizing across diverse patient anatomies and tumour characteristics, which are often not adequately captured in training models.

In comparison, Stephens et al.~\cite{stephens2024reinforcement} enhanced intensity-modulated radiation therapy (IMRT) by applying RL to reduce radiation doses to vital organs, specifically achieving a 7 Gy reduction in parotid gland doses. Their results are significant, but the simplified assumptions about fixed planning objectives may limit its applicability to complex clinical cases where multiple organs need simultaneous protection.

Further refining this, Purwati et al.~\cite{purwati2025analysis} applied RL and Monte Carlo simulations to optimize brachytherapy for right lung cancer, considering patient age variations. This approach highlights the flexibility of RL in personalized treatment but may face difficulties in real clinical settings where the precise anatomical and physiological parameters vary widely among patients.

Finally, Ferrante et al.~\cite{ferrante2025transforming} introduced an RL framework incorporating multimodal models and few-shot learning, aiming to enhance treatment planning. While innovative, this model still requires validation in clinical workflows, as integrating complex models into existing systems remains a significant challenge. While each study demonstrates RL’s potential to optimize radiotherapy, challenges such as data heterogeneity, computational demands, and real-world validation persist. These studies highlight the promise of RL in personalizing lung cancer treatment but underscore the need for further refinement and validation to ensure successful clinical integration. Appendix B Table~\ref{tab:Summary3} provides a summary of studies that have applied RL to personalized therapy, highlighting key approaches and findings.

\subsection{Disease Prediction and Diagnosis}
This section explores the application of machine learning and RL in disease prediction and diagnosis, divided into four principal categories: Primary Diagnosis Support, Diagnostic Imaging and Screening, Pathology and Rare Conditions, and Mental and Neurological Disorders. Each section addresses the relevant applications, techniques, results, and issues within various healthcare settings.

\subsubsection{Primary Diagnosis Support}
The primary diagnosis support employs RL to facilitate early disease identification and risk evaluation. This encompasses symptom checkers, triage systems, and preliminary risk assessment models that assist healthcare providers in making informed decisions more quickly and accurately.

\noindent\textbf{Symptom Checkers:} Symptom Checker 1.0, developed by Tang et al.~\cite{Tang2016}, is an RL-based symptom-checking system using 11 DQN models, each representing a different body part. The system, where physicians ask questions and diagnose based on patient responses, trained each model independently with symptoms encoded as one-hot states. Actions were rewarded based on accurate diagnoses and fewer inquiries, mimicking clinicians' behavior.

The system lacks real-time adaptability and contextual understanding, essential for accurate diagnoses in diverse clinical settings. Symptom Checker 2.0, developed by Kao et al.~\cite{Kao2018}, consists of a hierarchical RL approach with a master agent coordinating anatomical models for symptom checking. This system, using DQN at both levels, mimicked a group of doctors with varying expertise, improving diagnostic accuracy. The method outperformed traditional systems in self-diagnosis of health conditions. The system still faces challenges in handling diverse patient data and ensuring real-world applicability.

\noindent\textbf{Triage systems:} This section discusses the application of RL and clinical decision support systems (CDSS) in facilitating primary diagnosis for triage purposes. These technologies demonstrate the potential to improve triage efficiency and accuracy. However, obstacles persist. 

Kim et al. created a system employing wearable devices for mass casualty triage, attaining an AUC of 0.88 and an accuracy of 74.9\%. Nonetheless, issues regarding data variability and sensor precision were observed in practical applications~\cite{kim2021automated}. Abe et al., concentrated on the identification of severe traumatic intracranial haemorrhage detection, achieving an accuracy of 80\% with XGBoost, although enhancements are required for managing complicated data~\cite{Abe2022}.

Matta et al. assessed a Clinical Decision Support System (CDSS) in an emergency department, attaining 85\% sensitivity for urgent situations. However, its efficacy for nonurgent cases was suboptimal~\cite{maatta2023diagnostic}. Han et al. presented a multi-agent system utilizing a large language model (LLM) for triage, demonstrating the potential for resource allocation while encountering difficulties in high-pressure scenarios~\cite{han2025development}. The studies illustrate the capacity of RL to enhance triage decision-making. Although these systems exhibit potential in clinical environments, they continue encountering obstacles with data quality, flexibility, and efficacy in non-urgent situations. Additional development and real-time validation are essential for wider implementation.

\noindent\subsubsection{Early Risk Scoring:} RL is progressively utilized in healthcare to enhance early risk assessment, including the prediction of psychosis risk and the optimization of screening protocols. This subsection compares two studies employing RL for early risk assessment, one centered on breast cancer screening. Yala et al. utilized RL to enhance breast cancer screening, achieving a 10\% improvement in early detection rates and a 15\% reduction in the number of screens through a tailored methodology~\cite{yala2022optimizing}. The study depended on retrospective data from a singular institution, constraining its real-world applicability and generalizability. 

Luther et al.~\cite{luther2025probabalistic} used RL to predict psychosis risk in high-risk youth, finding poor reward learning linked to 20\% higher conversion and more severe negative symptoms. 

The~\cite{luther2025probabalistic} study, albeit enlightening, has a limited sample size of 46 CHR kids, constraining its generalizability. Both studies emphasize the potential of RL in healthcare while highlighting difficulties, including limited sample sizes and the necessity for real-world validation to improve generalizability and efficacy.

\subsubsection{Diagnostic Imaging and Screening}
Unlike predictive classifiers, RL-based diagnostic agents can evolve decision policies that adapt to patient feedback or imaging context over time, exemplifying the paradigm shift toward action-oriented learning. Recent breakthroughs in RL have significantly improved diagnostic imaging and screening, especially in identifying coronary artery disease (CAD), myocardial infarctions, cancer detection, and arrhythmias. Ghasemi et al. illustrated the efficacy of offline RL in enhancing CAD treatment, resulting in a 32\% enhancement in clinical outcomes. The study identified critical challenges, including data preprocessing and model adaptability, that hinder the scalability and generalization of RL across varied patient populations in real-world contexts~\cite{Ghasemi2025}. 

Similarly, Pradeep et al. integrated wavelet transformations with RL for the early identification of myocardial infarction, achieving a prediction accuracy of 86.04\%. Despite demonstrating considerable potential, challenges such as data imbalance and model assumptions undermined the robustness of the findings, restricting its applicability to wider clinical contexts~\cite{pradeep2025predicting}.

In the area of cancer detection, Wang et al. presented EarlyStop-RL, which minimizes false positives and diagnostic delays in lung cancer screening. This model exhibited notable enhancements but emphasized the necessity for real-world validation and clinical integration to ascertain its practical applicability across many healthcare contexts, a persistent problem for RL models in clinical environments~\cite{Wang2024}.\\
Qian Xiao et al. utilized RL for the diagnosis of myocarditis, attaining an F-measure of 88.2\% and a geometric mean of 90.6\%. Despite the encouraging results, the study recognized data imbalance and model assumptions as significant limitations, highlighting the necessity for larger representative datasets to assure generalizability and wider clinical applicability~\cite{Kasmaee2024}.\\
Serhani et al. concentrated on optimizing CNN hyperparameters for arrhythmia prediction, with an accuracy of 97.4\%. Notwithstanding this elevated accuracy, issues about real-time adaptation and data variability persist, hindering the practical implementation of RL models in dynamic healthcare settings~\cite{serhani2025enhancing}.\\
Tiwari et al. illustrated the potential of RL for automated disease detection, enhancing classification accuracy by 8.04\%. Although this enhancement is significant, the study highlighted the necessity for empirical validation and model integration, underscoring that these measures are essential for transitioning RL models from research environments to clinical applications~\cite{Tiwari2022}.\\
Finally, Yang et al. enhanced LLM alignment with radiologists for chest CT report summaries, resulting in a 2.56\% improvement in precision and a 1.77\% enhancement in recall. The study identified the misalignment between LLMS and radiologists as a critical obstacle to complete integration, emphasizing the necessity of addressing these problems for the extensive use of RL models in clinical workflows~\cite{Yang2025}.\\
Recent advancements in RL have significantly improved diagnostic accuracy across various other diseases. As evidenced by studies such as Takahashi et al.\cite{Takahashi2023} achieved 94.2\% precision and 95\% sensitivity for NASH/NAFLD diagnosis, while Chen et al.~\cite{Chen2022} attained 98.31\% accuracy in COVID-19 CT scan diagnosis. Tchango et al.\cite{Tchango2022} developed an RL-based differential diagnosis system, and Rashidi et al.\cite{Rashidi2025} enhanced diagnostic reliability and objectivity in liver disease. Nath et al.\cite{Nath2022} automated diabetic retinopathy treatment, showcasing RL’s potential in clinical automation.\\
These research studies collectively illustrate the transformative influence of RL on diagnostics and medical imaging. However, they encounter common problems, including real-world validation, data imbalance, and clinical integration. These deficiencies highlight the necessity for additional research to enhance RL applications in healthcare, focusing on scalability, adaptability, and clinical acceptance. 
\subsubsection{Pathology and Rare Conditions} RL has emerged as a transformative instrument in pathological diagnostics, improving the accuracy, efficiency, and predictive capabilities of various medical conditions. Syrykh et al.,~\cite{syrykh2025role} illustrate how AI, specifically deep learning and RL, aids pathologists in analyzing intricate datasets in the field of haematolymphoid diagnostics, thereby decreasing the likelihood of misdiagnosis. Nevertheless, challenges persist, notably in dataset reliability and the necessity of expert validation in real-world clinical settings, despite the potential. Similarly, Togootogtokh et al.,~\cite{togootogtokh2025voicegrpo} investigate the feasibility of applying RL to detect vocal pathology, resulting in remarkable diagnostic enhancements. Their model, which is founded on a mixture of expert transformer architecture and Group Relative Policy optimization (GRPO), significantly improves diagnostic accuracy. However, the dependence on synthetic datasets underscores the persistent data scarcity problem.\\
 These developments in AI and RL are not restricted to intricate datasets; they also encompass disease-specific applications, such as prostate cancer diagnostics, where Zhu et al.,~\cite{zhu2022artificial} implemented machine-learning models to aid pathologists in the diagnosis and grading of the disease. The study demonstrates that AI can enhance diagnostic precision; however, integrating such tools into clinical workflows is impeded by the necessity for more diverse, high-quality datasets and the ability to interpret models. CASANDE, a model that generates differential diagnoses and prioritizes severe pathologies, is a model that replicates the reasoning processes of doctors. This sentiment is echoed by Fansi Tchango et al.,~\cite{fansi2022towards}. Their work serves to emphasize the potential of RL to enhance diagnostics; however, the obstacle remains in the development of systems that can be seamlessly integrated into clinical practice while also being interpretable and trusted by healthcare professionals.\\
 Finally, Zheng et al.,~\cite{zheng2023learning} introduced FastMDP-RL, an RL framework for melanoma histopathology that enhances inference speed while preserving diagnostic accuracy. Nevertheless, this method, like others, also faces the challenge of balancing diagnostic precision and computational efficiency, particularly when working with high-resolution, large-scale data.\\
 These studies collectively emphasize the increasing potential of RL to transform pathological diagnostics. Although AI and RL are making substantial progress in enhancing diagnostic accuracy and efficiency, critical barriers to complete clinical adoption include data scarcity, model interpretability, and integration into clinical settings. 
\subsubsection{Mental disorder and seizures diagnostics} 
Mental health diagnostics have been transformed by RL, which has introduced novel approaches to understanding and forecasting disease progression. As RL is feedback-based, it is particularly useful for diseases like Alzheimer's, Parkinson's, Seizures and psychosis, where conventional methods often fail to address the intricacy and variability of symptoms.\\
A multitude of research has utilized RL in mental health diagnoses, each offering distinct methodologies and results while encountering similar problems. Yuan et al.\cite{yuan2025integrating} investigated the application of RL robots in dementia care, demonstrating that RL facilitated patient engagement through adaptive, context-sensitive interactions. This method emphasized the potential of RL in patient involvement; nevertheless, it was predominantly dependent on simulated environments, constraining its practical usefulness. Conversely, Saboo et al. employed RL to forecast the evolution of Alzheimer’s disease, attaining notable enhancements in long-term cognitive predictions compared to conventional approaches. Their model forecasted 10-year cognitive outcomes, illustrating RL's predictive capability; yet, data scarcity continues to be a significant obstacle for widespread application~\cite{saboo2021reinforcement}.\\
Bhattarai et al. adopted a treatment-focused methodology, employing RL to enhance treatment protocols for Alzheimer's patients with comorbidities such as hypertension, diabetes and chronic liver or kidney diseases. Although their model surpassed clinician-generated regimens in extensive datasets, it had difficulties with smaller ones. This underscores a critical constraint of RL models: dataset size, which can substantially influence performance in practical applications~\cite{bhattarai2023using}. Luther et al. predicted psychosis risk with probabilistic RL, accurately forecasting the transition to psychosis in high-risk adolescents. The study's limited sample size of 46 participants raises issues over the generalizability of the model's findings, particularly in comparison to bigger datasets utilized in previous research~\cite{luther2025probabilistic}.

Zhu et al. presented a Rule-based RL Layer (RRLL) in seizure detection, integrating domain-specific rules to enhance predictive accuracy. Although this method enhanced the model's physiological precision, the necessity for additional validation in practical clinical environments continues to hinder its broad implementation~\cite{zhu2025towards}. Hatami et al. employed RL for MRI classification in Alzheimer's diagnosis, enhancing accuracy and precision compared to conventional approaches. They emphasized that data imbalance is a continual problem that impacts the model's generalizability~\cite{hatami2024investigating}.

Zhao et al. utilized RL to adjust deep brain stimulation parameters for Parkinson's disease, enhancing symptom control and energy efficiency. Notwithstanding these encouraging outcomes, the model's dependence on real-time input highlights the difficulty of real-time adaptation, hence complicating its clinical use~\cite{zhao2025reinforcement}. 

As a result of these investigations, it has been demonstrated that RL can offer tailored therapy approaches, improve diagnostic accuracy, and address challenges such as data imbalance and modeling disease progression that is challenging in mental healthcare diagnostics.
\subsection{Surgical and Robotic Assistance} RL-powered robotic surgery improves precision, adaptability, and patient care by automating tasks like suturing and tissue manipulation. Nakao et al. showed robotic assistance in the double-flap technique reduced complications like stenosis and reflux esophagitis, highlighting RL's impact on improving patient outcomes through better surgical precision~\cite{nakao2025efficacy}. Similarly, Wang et al. proposed a standardized curriculum for robotic hepatic-pancreaticobiliary surgery, ensuring patient safety during the learning curve for novice surgeons. Although their methodology enhances safety, it mostly focuses on training rather than real-time surgical support~\cite{wang2025scoping}. Another significant advancement comes from Chen et al., who introduced an online trajectory guidance system using imitation learning and feedback to help novices improve their skills in real-time, substantially narrowing the disparity in manipulation skills between beginner and expert surgeons (76.47\% enhancement in task performance)~\cite{chen2025online}. 

Fan et al. developed a sim-to-real framework for automating tissue manipulation, reducing human intervention~\cite{fan2023sim}. In contrast, Fan et al. introduced Safe Experience Reshaping (SER), a framework to ensure safety during robotic surgery training. However, real-time implementation continues to cause issues~\cite{fan2024safe}. Lastly, Zargarzadeh et al. integrated robotic surgery with automatic blood suction, improving decision-making in complex surgeries like bleeding~\cite{zargarzadeh2025from}. 

These advancements show RL's potential to enhance robotic surgery by improving functionality and reducing human intervention. However, challenges persist in ensuring safety, adapting simulations to real-world scenarios, transferring expert skills to novices, and balancing automation with patient safety. 

Appendix B Table~\ref{Tale-4_summary} summarises significant studies utilising RL in many clinical applications, including diagnostics, triage systems, mental disorder assessment, and surgical task automation.
\subsection{Operational and Resource optimization in Healthcare: RL-driven Workflow Automation and Resource Allocation}
The integration of RL in healthcare, beyond the operating room, has shown great promise in optimizing various operational aspects, such as workflow automation and resource management. Several studies have highlighted this potential: 

Abdullah Lakhan et al.~\cite{lakhan2024mobile} explored combining blockchain technology and RL to optimize task outsourcing in mobile-edge computing (MEC) frameworks. This integration ensures data security, automates resource allocation, and addresses challenges related to data privacy and latency in environments like the Internet of Medical Things (IoMT). 

Abeer Z. Al-Marridi et al.~\cite{al2024optimized} extended this by introducing a multi-agent RL (MARL) framework to optimize resource allocation and secure data management. Their research employs blockchain technology to enhance privacy protection during medical data outsourcing. It dynamically adjusts resource allocation to meet fluctuating patient demands, but was restricted by the scalability and adaptability of the blockchain solution. On the other hand, Qiang He et al.~\cite{he2023blockchain} utilized deep RL (DRL) for secure data offloading in edge computing environments. Their system optimizes data outsourcing strategies, reducing operational costs while improving healthcare providers' access to critical medical information. Concerns regarding the energy consumption and latency costs associated with blockchain integration persist. Teddy Lazebnik et al.~\cite{lazebnik2023data} focused on hospital resource allocation and staffing optimization using RL and agent-based simulation. By training an RL agent with real-world data, their approach helps automate staffing decisions and resource distribution, improving operational efficiency and service delivery in hospitals. Lastly, Mauricio Tec et al.~\cite{tec2025rule} introduced Rule-Bottleneck RL (RBRL), combining RL with large language models (LLMs) to enhance decision transparency. RBRL ensures that AI-driven resource allocation decisions in hospitals are justifiable and understandable by healthcare professionals while improving resource management. However, practical implementation in varied healthcare settings and with non-standardized data continues to be difficult. 

These findings underscore the potential of RL to enhance hospital processes, data management, and resource allocation. Nonetheless, issues such as scalability, real-time adaptation, and blockchain integration for safe data management endure. Additional research is required to enhance these models and tackle the issues associated with real-world implementation in healthcare systems.

\section{Clinical-Grade RL: From Simulation to Trustworthy Deployment}
RL introduces powerful capabilities for sequential decision-making, but its deployment in healthcare must navigate a complex mix of ethical, technical, and operational challenges. Unlike supervised learning models that passively predict outcomes, RL agents actively recommend or practice treatment policies. Therefore, healthcare RL solutions carries a risk of increased consequences with any model misalignment, poor interpretability, or unsafe exploration. This section surveys a range of challenges from algorithmic design to broader considerations of deployment, responsibility, and trust in clinical environments.

\subsection{Reward Misspecification and Clinical Alignment}
In RL, determining what to optimize is often challenging than optimizing it. Clinical goals are transferred into algorithmic goals through the use of reward functions, making reward design one of the most ethically and technically sensitive components of RL in healthcare. While long-term outcomes such as 90-day mortality or hospital readmission have been commonly used in literature~\cite{wan2024multiobjective}, these proxy functions are often sparse, delayed, and influenced by factors external to clinical care (e.g., socioeconomic status). They also fail to capture short-term physiological goals such as organ support or patient comfort. Intermediate rewards like SOFA scores or TTR offered more immediate feedback, but they too are imperfect proxies, that may not capture the true long term objective for patient care. Such misaligned reward designs can lead to irrational incentives such as optimizing for early discharge at the cost of patient safety~\cite{wan2024multiobjective}.

As a solution recent advances in reward learning introduced techniques such as Inverse Reinforcement Learning (IRL)~\cite{deshpande2025advances} and Bayesian IRL~\cite{ramachandran2007bayesian} that attempt to infer latent clinical preferences from expert demonstrations. However these methods rely heavily on the assumption that expert behavior is optimal. Causal RL frameworks~\cite{du2024situation} incorporate clinical knowledge through structural causal models (SCMs)~\cite{lu2022efficient}, enabling counterfactual reasoning, but they require detailed domain expertise and remain hard to validate empirically.

As RL systems increasingly influence real-world decisions, there is a growing call to move beyond proxy functions and co-design reward functions with clinical experts in the loop from early design stages. Such collaborative approaches could develop reward functions with safety constraints, multi-objective trade-offs, and value-sensitive outcomes from the outset.

\subsection{Sparse, Static Data and Imperfect Demonstrations}
Most healthcare RL systems are trained offline on retrospective data, which poses a fundamental risk of misalignment with RL’s interactive foundations. Furthermore, static datasets suffer from limited coverage in the state-action space, leading to out-of-distribution (OOD) actions during policy execution. These OOD actions are known to produce inflated Q-values and unsafe recommendations~\cite{prashant2025guaranteeing}. While techniques such as rare action removal reduces this risk, it also risks discarding valuable clinical interventions.

To address this, conservative algorithms such as BCQ, CQL, BRAC, and IQL were introduced. BCQ restricts actions to dataset distributions using a VAE, while CQL penalizes high Q-values for unseen actions. BRAC aligns learned and behavior policies using divergence measures~\cite{zhang2024implicit}, and IQL avoids explicit policy improvement altogether, focusing on advantage-weighted regression. The comparative analysis of these methods are illustrated in Table 7.

Recently introduced hybrid methods such as WD3QN incorporate clinician priors to anchor action selection to expert values in critical states to offer a safeguard against unsafe extrapolations. IRL based methods also tend to suffer under noisy and inconsistent demonstrations. Approaches like D-REX~\cite{chen2021learning} introduce noise-aware ranking, and Bayesian IRL captures uncertainty in inferred rewards, improving robustness in ICU environments.

Model-based RL augment limited data by simulating counterfactual trajectories. However, the notorious compounding model error often degrades performance in complex clinical settings. Causal RL attempts to overcome this by learning SCMs that clarify treatment-outcome relationships and correct for confounding~\cite{lu2022efficient}, but these methods remain fairly new and underdeveloped in the healthcare setting.

\subsection{Partial Observability and Belief Estimation}
Typically, healthcare settings are partially observable with noisy, missing, or delayed information. This violates the critical MDP assumptions and requires agents to infer latent states from incomplete observation histories. POMDP-based models and belief-state estimators such as Deep Variational RL (DVRL)~\cite{igl2018deep} and recurrent architectures~\cite{kosarkar2025design} have shown potential to overcome these issues. However they suffer from training stability and interpretability, where their performance is often highly sensitive to data sparsity and temporal resolution. 

\subsection{Data Sharing, Privacy, and Equity Gaps}
Deploying RL solutions in real-world is also constrained by data access and generalization. Patient data is often restricted, heterogeneous, and regulated under privacy laws like HIPAA and GDPR~\cite{rieke2020future}. Federated RL (FRL) provides a soltuio by enabling decentralized learning across institutions while preserving patient confidentiality~\cite{pei2024review, fang2025provably}. However, challenges remain in handling client heterogeneity, communication overhead, and convergence across distributed environments.

Training RL algorithms using biased data with under representation of marginalized populations can heighten health disparities when encoded into RL policies. Ensuring equitable treatment recommendations requires not just technical fixes, but proactive efforts in dataset curation, policy evaluation, and stakeholder engagement.

\subsection{Explainability and Clinical Trust}
Explainability in the learned policies and processes is key requirement For RL to successfully transition from lab settings to clinical deployment. Unlike supervised models that classify or regress, RL agents generate sequential decisions making it harder to attribute specific actions to interpretable causes. Traditional techniques such as SHAP and LIME are limited in capturing long-term dependencies or decision rationales in trajectory space~\cite{xu2024xrl}.

Emerging methods in explainable RL (XRL) aim to generate trajectory-level summaries, identify salient decision points, or cluster patient states by policy behavior~\cite{cheng2025survey}. However, many remain designed for synthetic environments, and few translate well to tabular, noisy clinical data.

Clinician-in-the-loop interfaces and human-centered evaluation of explanation quality measured using metrics such as cognitive load or decision confidence would encourage safe, acceptable RL adoption in healthcare.

\subsection{Policy Evaluation and Off-Policy Metrics}
At this point, online deployment of RL for real-time evaluation in healthcare is rarely feasible due to high patient risk. Instead, Off-Policy Evaluation (OPE) techniques are used to assess learned policies against historical data. These include Direct Methods (DM), Importance Sampling (IS), Doubly Robust (DR), and IS variants such as CWPDIS~\cite{luo2024position}. Table~\ref{tab:off_policy_comparison} illustrates some of these commonly used OPE techniques used to quantitatively evaluate DTRs. Each method balances bias and variance differently, and the selection depends on factors such as available data, policy divergence, and horizon length. 

Moreover, these OPE metrics are often supplemented by clinical plausibility checks such as comparing mortality correlations with Q-values, or using Kaplan-Meier curves to compare policy adherence. However, evaluation across studies remains largely inconsistent, urgently requiring a standardized benchmarking framework for OPE in healthcare RL.

\subsection{Fusion pitfalls in clinical RL}
Various fusion levels exposes failure modes that recur in clinical RL and matter for safety. First, multi-rate monitors, labs, and notes can produce misleading states unless encoders explicitly handle irregular sampling and changing distributions (e.g., GP/RNN state models and partial-observability–aware objectives)~\cite{xu2025meddreamer}. Moreover conflicts within sources could occur, and treating control as a POMDP with belief tracking and propagated uncertainty has proven practical in ICU dosing~\cite{futoma2020popcorn}. Cross-site bias is common where policies trained at one hospital degrade elsewhere; federated/offline RL offers a remedy by learning across sites without sharing raw data and has been demonstrated on multi-hospital sepsis cohorts~\cite{zhou2024federated}. Finally, decision-fusion governance is critical, and embedding clinical expertise or oversight into the policy (or evaluation) improves safety and acceptability in practice, as shown in sepsis treatment and ICU sedation dosing~\cite{wu2023value}.
\section{Design Dilemmas: Navigating Trade-offs in Healthcare RL}
As RL aims to shift the healthcare AI paradigm from passive inference to active decision-making, it introduces fundamental trade-offs in algorithm design and deployment. In healthcare, these trade-offs are heightened by factors such as limitations in quality and quantity in data, safety constraints, and the need for interpretability. This section explores the implicit assumptions and competing objectives that shape RL's performance and trustworthiness in clinical environments, that needs to be carefully handled for confident deployment in real world.
\subsection{Sample Efficiency vs. Generalization}
Popular on-policy algorithms such as PPO are known to be data-hungry, but provide stable policy updates. in contrast, off-policy methods such as DQN and DDPG) are more sample-efficient but tend to suffer from divergence issues in offline settings~\cite{valencia2025ctd4, luo2024serl}. Actor-critic were introduced to balance these extremes but still require careful tuning. In simulation-rich settings such as symptom checkers, on-policy RL would be more suitable, whereas in high-stakes or low-data domains, conservative off-policy learning is preferred.
\subsection{Optimization Stability vs. Expressiveness}
Value-based methods optimize Bellman error but tend to converge to suboptimal or unstable policies in highly complex environments such as healthcare~\cite{obando2024value}. In contrast, Policy gradient methods converge more reliably but often plateau at local optima. Model-based methods reduce sample use by simulating trajectories, but model misspecification lead to cascading errors~\cite{zhang2024optimizing}.
\subsection{Episodic vs. Infinite Horizon Framing}
Typically, clinical RL tasks are episodic with terminal outcomes such as mortality (diagnosis) in sepsis management (symptom checkers)~\cite{liang2024episodic, li2025safe}. However, chronic care requires infinite horizon modeling, where stable long-term policies are needed. This choice affects RL discounting, evaluation, and even feasibility of off-policy learning~\cite{harada2024longitudinal, hammoud2024evaluating}.
\subsection{Exploration vs. Safety}
Exploration is essential in RL to learn policies with sufficient coverage, but it can also lead to unsafe healthcare policies. Therefore, Randomized strategies like $\epsilon$-greedy or Thompson sampling could be used to constrain the exploration to avoid risky actions. Additionally, techniques such as conservative exploration and hopeful exploration~\cite{mohamed2015variational, johnson2016malmo, houthooft2016vime} provide safer alternatives, but lead to inefficient learning. This shows a key trade-off of balancing learning progress against ethical risk in clinical RL.
\subsection{Reward Optimization vs. Human Acceptability}
Moreover, a technically optimal policy may not always align with what is ethically or clinically acceptable, especially due to misalignment in reward design and actual clinical goals. For example, optimizing for mortality alone may neglect patient comfort or treatment equity. Multi-objective RL and human-in-the-loop practices are needed to alleviate this gap, with better reward shaping tools, clinician collaboration, and robust preference elicitation.
\subsection{Fusion trade-offs} 
Rich multi-modal fusion improves observability but increases sample complexity and OPE variance; simpler feature sets are stable but risk omitted-variable bias. Ensemble critics (decision-fusion) improve reliability yet can under-explore. Federated fusion enhances privacy but may reduce calibration across sites.

Overall, the success of RL in healthcare will depend not just on technical innovation, but on the careful negotiation of these trade-offs-reflecting the real-world complexity of patient care.

\begin{table*} 
\centering
\begin{small} 
\caption{Comparison of conservative RL Methods}
\scalebox{0.95}{
\begin{tabular}{|p{2.5cm}|p{3cm}|p{3cm}|p{3.5cm}|p{3.5cm}|}
\hline 
\rowcolor{headerpink}
\textbf{Feature / Method} & \textbf{IQL (Implicit Q-Learning)} & \textbf{CQL (Conservative Q-Learning)} & \textbf{BRAC (Behavior-Regularized Actor Critic)} & \textbf{BCQ (Batch-Constrained Q-learning)} \\ \hline
\textbf{Main Strategy} & Implicit policy learning via advantage weighting & Conservative Q-learning by penalizing Q(OOD actions) & Regularize policy to stay near behavior policy & Constrain action selection to dataset actions \\ \hline
\textbf{Handles Q-over estimation} & Yes (implicitly) & Yes (explicit penalty) & Yes (via regularization) & Yes (action constraints) \\ \hline
\textbf{OOD Action Handling} & Implicitly avoids via no policy improvement & Penalizes OOD Q-values & Regularizes toward behavior policy & Only evaluates in-distribution actions \\ \hline
\textbf{Requires Behavior Policy Estimation} & No & No & Yes (for regularization term) & Yes (generative model like VAE) \\ \hline
\textbf{Policy Type} & Deterministic, derived from weighted advantage & Explicit actor or argmax(Q) & Explicit actor (stochastic/deterministic) & Argmax over sampled dataset-like actions \\ \hline
\textbf{Uses Generative Model} & No & No & No (but needs behavior policy model) & Yes (VAE to model behavior actions) \\ \hline
\textbf{Training Stability} & Stable (no policy iteration) & Can be unstable (sensitive to penalty strength) & Depends on quality of behavior policy estimation & Stability depends on generative model quality \\ \hline
\textbf{Hyperparameter Sensitivity} & Low to Moderate & High (penalty weight critical) & Moderate (regularization weight tunable) & Moderate (VAE, sample count need tuning) \\ \hline
\textbf{Exploration / Exploitation Tradeoff} & Implicit via advantage weights & Conservative — prioritizes safety & Tunable via regularization strength & Conservative — chooses only dataset-like actions \\ \hline
\textbf{Best Use Case} & High-dimensional data, scalable, easy implementation & Safety-critical applications, reliable conservatism & When behavior policy can be modeled well & Small-scale tasks, reliable datasets, safety-critical apps \\ \hline
\end{tabular}
}
\end{small}
\end{table*}

\begin{table*} 
\centering
\begin{small} 
\caption{Comparison of Off-Policy Evaluation Methods}
\label{tab:off_policy_comparison}
\renewcommand{\arraystretch}{1.3}
\scalebox{0.95}{
\begin{tabular}{|p{3cm}|p{5cm}|p{4cm}|p{4cm}|}
\hline
\rowcolor{headerpink} \textbf{Method} & \textbf{Key Idea} & \textbf{Strengths} & \textbf{Weaknesses} \\
\hline
Direct Method (DM)\cite{luo2024position} & Learn a model of rewards or Q-values (e.g., via supervised learning), then estimate value using this model & Simple; no need for importance weights & High bias if the model is wrong; can’t account for unobserved data \\
\hline
Importance Sampling (IS)~\cite{luo2024position} & Weight returns based on probability ratio between target and behavior policy & Theoretically unbiased; doesn’t need reward/Q model & Very unstable; sensitive to small action probabilities \\
\hline
Weighted IS (WIS)~\cite{luo2024position} & Normalize IS weights to sum to 1 & Reduces variance; more stable than IS & Still unstable; bias introduced; weights still noisy \\
\hline
CWPDIS~\cite{luo2024position} & Per-step variant of WIS ; weights at each time step; handles time-varying policies & Better for long horizons; balances bias/variance well & Needs accurate per-step policy estimates; still sensitive to rare actions \\
\hline
Doubly Robust (DR)\cite{luo2024position} & Combine DM and IS: use model for base value, IS for correction & More accurate than DM/IS; robust to either model or weight errors & Needs both an accurate model and weights; unstable if both are poor \\
\hline
\end{tabular}
}
\end{small} 
\end{table*}

\section{Emerging Frontiers and Future Directions}
Future directions in RL research are not merely technical enhancements but foundational steps toward institutionalizing RL as a new paradigm in clinical intelligence, enabling trustworthy decision-making in healthcare. New and emerging RL methods such as multi-agent systems, federated RL, and generative modeling aim to bridge the gap between theoretical performance and actionable, scalable deployment. These approaches expand the frontier of prescriptive analytics extending RL’s capacity to influence outcomes across distributed, multi-stakeholder, and ethically constrained care environments. An overview of these key emerging trends is provided in Figure~\ref{F11} below, illustrating the diverse directions in which RL is being applied to address complex challenges.
\begin{figure*} 
 \centering
\includegraphics[scale= 0.53]{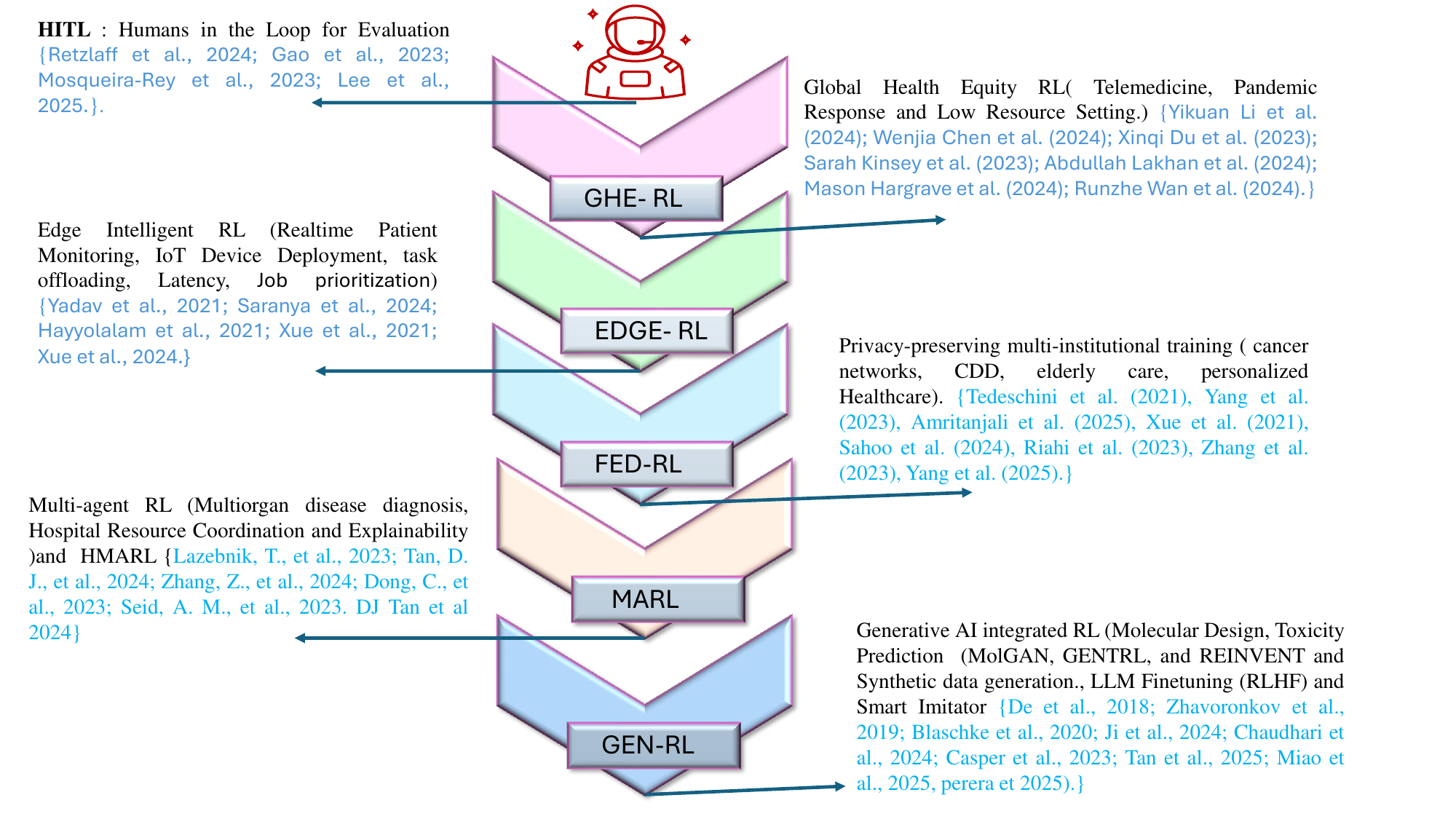} 
 \caption{Overview of Emerging Frontiers in RL}
 \label{F11}
 \end{figure*}
\subsection{GEN-RL}
Generative AI and RL have the potential to advance fields like molecular design and toxicity prediction significantly. While Generative AI excels at creating data, it struggles with optimization for specific objectives, a challenge RL can address. By combining the two, models can iteratively improve, targeting specific goals such as enhancing pharmacological activity and reducing toxicity, thus improving drug discovery efficiency. Models like MolGAN~\cite{de2018molgan}, GENTRL~\cite{zhavoronkov2019deep}, and REINVENT~\cite{blaschke2020reinvent} use RL to refine molecular outputs and enhance drug discovery accuracy. Apart from these, to improve the capacity of RL to manage imperfect clinical data, the Smart Imitator framework~\cite{perera2025smart} incorporates adversarial cooperative imitation learning. This approach has the potential to enhance personalized treatment policies and drug discovery. \\
However, challenges in reward modeling, including bias, reward hacking, and preference-based model limitations, persist. AI-based feedback and inverse RL (IRL) are being explored to address these issues. In healthcare, ensuring safety and ethical alignment in clinical recommendations is critical, and RL from Human Feedback (RLHF) has been instrumental in improving the safety of large language models (LLMs). Key contributions from Ji et al.~\cite{ji2024pku}, Chaudhari et al.~\cite{chaudhari2024rlhf}, Casper et al.~\cite{casper2023open}, and Tan et al.~\cite{tan2025equilibrate} have focused on safety alignment and balancing helpfulness with safety.\\
Despite improvements, challenges such as reward over-optimization, edge cases, and scalability remain. Future work should prioritize model robustness, resource efficiency, and continuous monitoring. The InfoRM framework~\cite{miao2025inform} is addressing reward over-optimization, enhancing model generalization, and improving the reliability of clinical recommendations.
\subsection{MARL} Multi-Agent RL (MARL) is revolutionizing healthcare by optimizing multi-hospital resource coordination, such as bed management and personnel distribution, ensuring efficient resource use in decentralized systems~\cite{lazebnik2023data}. In multi-organ disease detection, especially in complex cases like sepsis, MARL coordinates agents across organs to improve patient outcomes~\cite{tan2024advancing}. It also enhances decision-making transparency through techniques like decision-tree extraction, aiding healthcare practitioners in high-stakes contexts~\cite{zhang2024advancing}.\\
In IoMT and mobile edge computing, MARL improves task offloading by dynamically managing resources and enhancing service delivery, particularly in emergencies~\cite{he2023blockchain}. Its integration with Federated Learning (FL) enables privacy-preserving decentralized model training, addressing data security in real-time healthcare monitoring~\cite{zhang2024advancing}. Additionally, the MARL has been successfully implemented in managing multi-organ diseases, including sepsis, through frameworks such as HMARL~\cite{tan2024advancing}. This framework utilizes hierarchical agents to coordinate treatment strategies across organ systems, demonstrating how MARL can improve resource coordination and decision-making in healthcare.\\
Despite its success, MARL faces challenges in sample efficiency and explainability, which are critical for clinical trust. Future research should focus on improving these areas for better real-time performance in emergency healthcare contexts. 
\subsection{FED-RL} Federated RL (FED-RL) is gaining traction in healthcare for its ability to preserve data privacy while enabling collaborative, decentralized learning. It has been applied in clinical decision support systems (CDSS) for remote monitoring and geriatric care, offering real-time, privacy-preserving decision-making~\cite{Tedeschini2021}. FED-RL also enables tailored healthcare solutions like activity monitoring, fall detection for the elderly, and decentralized medical image analysis~\cite{Yang2023},~\cite{Amritanjali2025}. Additionally, it enhances the examination of Electronic Medical Records (EMRs) in decentralized systems, improving therapeutic outcomes~\cite{Tedeschini2021},~\cite{Xue2021}.\\
However, challenges persist, such as data heterogeneity, security threats like model inversion attacks, and high communication overhead in resource-limited environments~\cite{Sahoo2024},~\cite{Xue2021}. Non-IID characteristics of healthcare data can degrade global model performance, requiring strategies like tailored aggregation and adaptive model weighting~\cite{Sahoo2024}. Privacy concerns, particularly model inversion attacks, necessitate enhanced privacy-preserving methods like homomorphic encryption~\cite{Xue2021}.\\
Emerging methods like federated multi-agent RL (FED-MRL) and blockchain-based solutions aim to address these issues, improving scalability and security~\cite{Tedeschini2021},~\cite{Yang2025},~\cite{Riahi2023}. Tackling client attrition is also critical for maintaining system robustness~\cite{Sahoo2024},~\cite{Xue2021}. Overcoming these challenges will enable FED-RL to improve its application in healthcare, making systems more resilient, privacy-preserving, and scalable~\cite{Yang2025},~\cite{Zhang2023}.\\
\textbf{EDGE-RL} Edge intelligence is key to improving RL in healthcare by enabling real-time, efficient patient monitoring on Iot devices. Lightweight RL models help minimize latency, conserve energy, and ensure privacy through techniques like federated learning, making healthcare systems scalable and cost-effective. Studies have enhanced RL for healthcare, such as Yadav et al.'s CORL for optimizing latency and energy by offloading tasks to edge nodes~\cite{yadav2021computation}, Saranya et al.'s cloud-edge resource allocation with Deep RL and LOA~\cite{saranya2024enhanced}, and Hayyolalam et al.'s DRL for edge decision-making to reduce costs~\cite{hayyolalam2021edge}. Xue et al. explored federated RL for privacy-preserving clinical decisions~\cite{xue2021resource} and optimized energy and latency with DDPG-based RL offloading~\cite{xue2024low}, addressing data processing challenges like privacy and job prioritization. These studies highlight the role of edge-based RL in enhancing healthcare system efficiency, scalability, and privacy.\\
However, challenges remain, such as network congestion, slow performance due to federated learning, and the need for adaptive RL models across diverse edge devices. optimized offloading, privacy, and scalability solutions are essential to overcome these issues.\\
\subsection{GHE-RL} RL is advancing global health equity by improving healthcare in resource-limited areas. It optimizes resource allocation in pandemic response and telemedicine, enabling equitable healthcare access. Studies such as Yikuan Li et al.'s ventilator allocation~\cite{yadav2021computation}, Wenjia Chen et al.'s teleconsultation scheduling~\cite{chen2024teleconsultation}, and Xinqi Du et al.'s epidemic management~\cite{du2023epicare} contribute to this effort. Sarah Kinsey et al. used RL for diabetes prevention in rural areas~\cite{kinsey2023building}, while Runzhe Wan et al. applied it for infectious disease management~\cite{wan2024multiobjective}. Mason Hargrave et al. and Abdullah Lakhan et al. focused on enhancing treatment protocols and remote surgeries~\cite{hargrave2024epicare},~\cite{lakhan2024fiber}.\\
However, challenges like data scarcity, model biases, lack of interpretability, and scalability remain. Addressing these is crucial for RL's full potential in promoting health equity.\\
\textbf{HITL} Healthcare is exploring Human-in-the-Loop (HITL) RL despite established evaluation measures, which are already discussed in the above section. Model accuracy, reliability, and safety depend on clinician feedback during training. Retzlaff et al.~\cite{retzlaff2024human} emphasize the significance of human engagement in learning and evaluation, ensuring models meet therapeutic standards. A framework by Gao et al.~\cite{gao2023offpolicy} enhances the assessment of sparse human feedback in healthcare contexts through off-policy evaluation (OPE). According to Mosqueira-Rey et al.~\cite{mosqueira2023humanfeedback}, HITL methods such as active learning and machine teaching improve model performance by incorporating human supervision. Lee et al.~\cite{lee2025artificial} show how HITL in mental health uses patient feedback to tailor AI-driven therapy sessions to match patient needs.\\
These results emphasize the relevance of clinician input in refining RL models and improving healthcare outcomes. Future research should focus on more efficient feedback systems, model transparency, and real-time scalability.
\section{Discussion}
RL has emerged as one of the most promising and conceptually novel approaches used in healthcare AI. Its ability to effectively learn through interaction, optimize decisions over time, and adapt to changing patient states differentiates RL from conventional prediction-based models. RL has demonstrated the ability to improve healthcare across various aspects such as personalize treatment strategies, balance competing risks, improve long-term outcomes, and optimize hospital operations under uncertainty. These capabilities position RL not merely as a technical enhancement, but rather a paradigm shift toward prescriptive, policy-driven clinical intelligence.

However, despite the growing academic interest and substantial technical progress, real-world deployment of RL in healthcare is limited. The gap between promise and practice reflects deeper challenges that span methodological, infrastructural, and sociotechnical aspects. Through this survey, we identify several key underexplored yet essential dimensions to guide the field toward safe and meaningful clinical adoption.

\noindent\textbf{Beyond Prediction - The Comparative Value of RL:} Primary value in RL in healthcare lies in its ability to manage sequential decisions under uncertainty, which is commonly found in intensive care, oncology, and chronic disease management. Supervised or rule-based systems can perform well in settings with immediate feedback and limited action spaces. However RL excels in high-stakes, temporally extended environments. By continuously updating policies based on feedback, RL agents possess a level of personalization and adaptability that cannot be achieved through static models. This potential is already validated through several successful demonstrations. Specifically, RL models have optimized sedation and weaning protocols in ICUs, adjusted insulin dosing for type 1 diabetes, and supported sepsis treatment decisions using retrospective clinical data. These early results illustrate that RL systems can effectively recommend actions optimized for long-term outcomes, not just risk scores.

\noindent\textbf{Methodological Fit to Clinical Context:} RL field has matured over the years, and various RL methodologies were introduced to overcome the unique challenges encountered in different problem settings. The effectiveness of an healthcare RL solution highly depends on how well the selected methodologies align with the addressed clinical problem. For example, in ICUs, offline RL is often preferred to ensure patient safety while learning from historical data. Model-based RL is growing in popularity in interpretable domains such as radiotherapy planning or diabetes management, where transparency and clinical validation are paramount. In acute care, off-policy, conservative methods provide safety while preserving adaptability. For chronic conditions, RL should focus on long-term policy optimization, adapting treatments to patient trajectories over time. In diagnostic imaging and pathology, exploration-heavy RL techniques could uncover new patterns in complex, high-dimensional datasets. These examples highlight RL’s flexibility across use cases, while highlighting the need for rigorous design tailored to each clinical setting.

\noindent\textbf{Clinical Integration - From Algorithm to Action:} Unlike traditional prediction based models, RL agents require continuous interaction with the environment, including real-time feedback loops and policy updates. However, existing healthcare infrastructure is not inherently designed to cater this demand. Accordingly, integration into EHRs is constrained by interoperability issues, legacy systems, and workflow complexity. Furthermore, most RL implementations rely on structured tabular data, yet real-world care mostly involves multimodal signals from clinical notes and labs to imaging and wearable data. Future RL systems must be able to reason across modalities for holistic, context-aware decision-making.

\noindent\textbf{Generalizability and Equity:} Similar to other AI solutions, RL models trained on data from a single institution or region often suffer from poor generalization across different populations and care settings. This challenge is particularly severe in healthcare, where demographic diversity, infrastructure variation, and local practice norms could heavily influence final outcomes. However recent techniques such as federated learning, domain adaptation, and meta-learning show promise but remain underexplored in clinical RL contexts. Therefore effective RL adaptations with broader generalization are needed to avoid potential health disparities, and improve confidence in RL in healthcare.

\noindent\textbf{Accountability, Explainability, and Trust:} The efforts to make autonomous RL systems are challenged by complex questions about accountability, auditability, and interpretability. In high-stakes domains, clinicians and regulators must be able to understand what a model recommends and more importantly why it is recommended. Poorly aligned reward functions, black-box policies, and non-intuitive behavior harm clinician trust and regulatory approval. Ensuring explainable and human-in-the-loop RL systems is essential for responsible clinical adoption.

\noindent\textbf{Lessons from Other Domains:} The promise of RL in healthcare is further validated through the success from various adjacent fields. AlphaZero showcases optimal policy learning in uncertain, adversarial environments, which are analogous to taking high-risk medical decisions~\cite{silver2017mastering}. Tesla’s autopilot demonstrates continuous adaptation in dynamic real-world settings~\cite{dikmen2016autonomous}. Recent trends in DeepSeek and other large language model systems showcase how RL can handle open-ended reasoning and domain transfer~\cite{liu2024deepseek}. IBM Watson Health illustrates the power of ingesting large, multimodal datasets for real-time decision support~\cite{yang2020rise}. These examples ensures RL’s core strengths in sequential optimization, adaptability, and long-horizon reasoning—map directly onto healthcare’s needs.

\noindent\textbf{Persistent Limitations and Future Directions:} Despite these advances, several core limitations still remains. Many RL models still rely on oversimplified rewards such as in-hospital mortality, ignoring more refined and practical outcomes like long-term function, cost, or quality of life. Current solutions also lack prospective validation raises questions about real-world performance. Most systems are trained on retrospective, static data, risking overfitting and unsafe generalization. Furthermore, in domains like robotic surgery, poorly defined incentives could result in unsafe or unethical actions. Finally, overcoming these persistent limitations is paramount and requires rigorous evaluation pipelines including simulation environments, synthetic data augmentation, real-world pilot studies, and continuous monitoring.

However emerging advances in RL show great promise in addressing these limitations. For example, Batch Constrained Q-Learning (BCQ) and Weighted Duelling Double Q-Networks (WD3QNE) have demonstrated improved safety in high-risk clinical contexts. Generative RL (Gen-RL) could support personalized policy generation by simulating diverse clinical trajectories. Federated RL (Fed-RL) offers a pathway to learn from distributed data while preserving privacy and institutional control. The integration of RL with generative AI, multimodal representation learning, and causal inference frameworks could further enhance performance, robustness, and transparency. However, these needs to be attemtped with caution to avoid over hype and misalignment with clinical values.

Finally, to fulfill its transformative potential, RL should evolve beyond benchmark-driven development and into clinically robust, ethically grounded, and context-aware systems. The shift from prediction to action is not just technical, it is institutional, ethical, and collaborative. Future progress will require interdisciplinary partnerships, transparent governance, and a sustained focus on real-world readiness for eventual success and wide spread application. If achieved, RL can become not only a symbol of healtcare AI culture, but a cornerstone of next-generation clinical intelligence, that is intelligent, adaptive, and above all, trustworthy.
\section{Conclusion}
RL holds transformative potential in healthcare AI from retrospective prediction to real-time, adaptive decision optimization. By enabling agents to learn from interaction and optimize for long-term outcomes, RL offers a powerful framework for transforming how clinical and operational decisions are made in dynamic, complex environments. This survey systematically reviewed the healthcare RL landscape, providing a structured taxonomy of methodologies, including model-free, model-based, and offline approaches. Furthermore we critically examined their deployment across diverse domains varying from drug dosing, surgical robotics, ICU management to diagnostics. Framed through information fusion, clinical RL becomes a principled machinery for integrating multi-sensor evidence, multi-institutional knowledge, and multi-objective goals into safe, adaptive policies.

While notable progress has been made since 2020, real-world adoption remains constrained by persistent challenges such as data sparsity, biased or poorly aligned reward functions, limited prospective validation, generalizability gaps, and the lack of explainable policy representations. We extended the review on technical innovations such as conservative off-policy learning, policy regularization, and privacy-preserving collaboration that aim to address these barriers. We also highlighted the emergence of frameworks such as Generative RL, Federated RL, and multi-agent RL, that holds a great promise to further enhance personalization, privacy, and scalability in complex real-world healthcare settings.

A key message of this review is that advancing RL in healthcare is not solely a technical endeavor. Achieving trusted clinical adoption requires interdisciplinary collaboration across AI research, clinical practice, ethics, and regulatory governance. Realizing RL’s full potential demands systems that are not only intelligent and optimal under standard performance metrics, but also interpretable, safe, and aligned with patient and provider values. As healthcare is reaching a key transformative moment in the adoption of autonomous decision systems, RL offers more than incremental improvement. It provides a pathway toward intelligent, adaptive, and personalized systems that continuously learn, evolve, and act to optimize health outcomes in the face of uncertainty.

We hope this survey serves as both a foundation and a forward looking roadmap, inspiring new research, informing safe translation, and guiding the responsible integration of RL into real-world healthcare applications.

\printcredits

\bibliographystyle{cas-model2-names}
\bibliography{els-cas-templates/Final_paper_09Aug}

\onecolumn 
\textbf{Appendix A}
\begin{figure}[!h]
 \centering
 \includegraphics[scale=0.45]{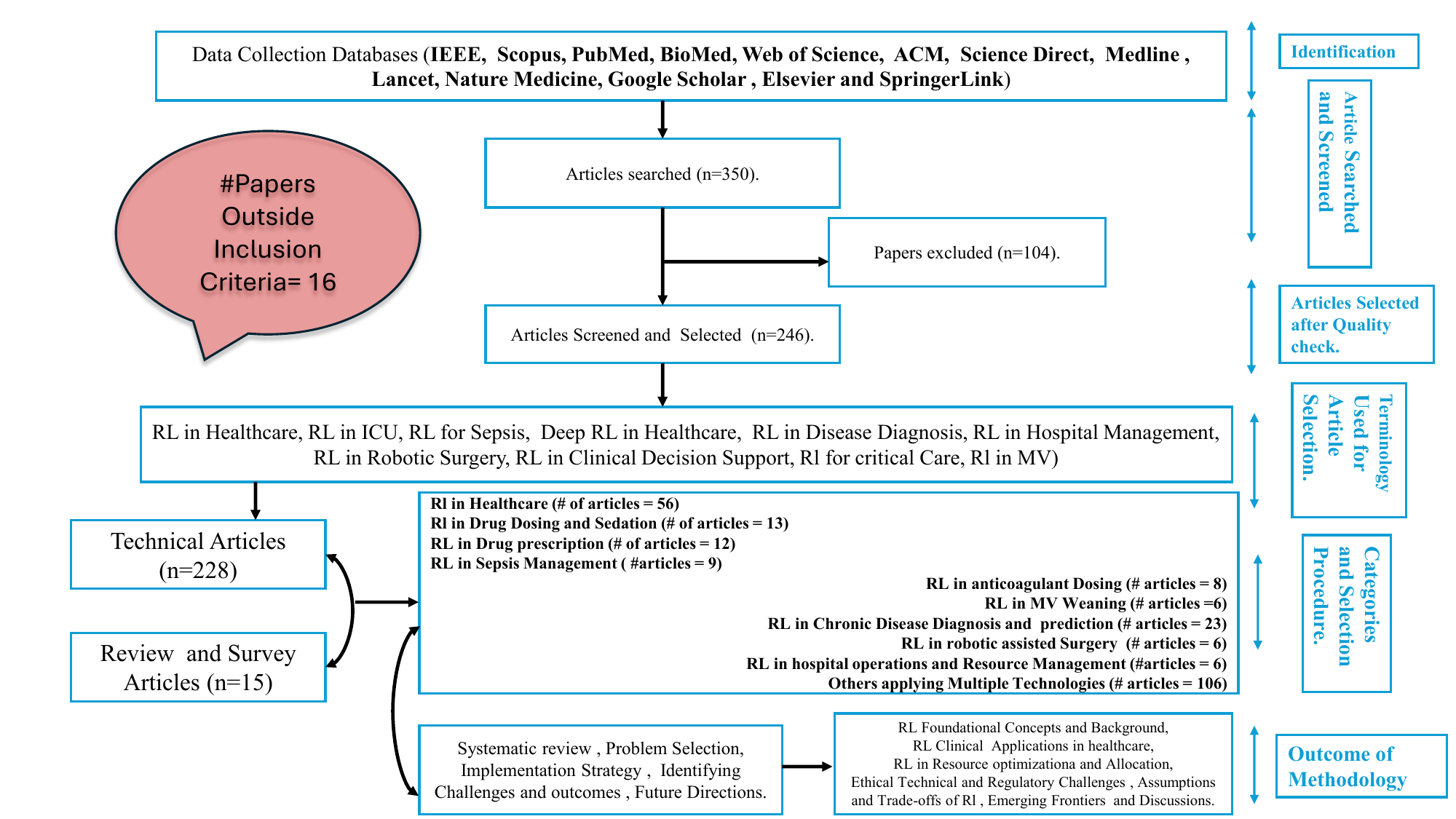} 
 \caption{PRISMA Framework: A structured approach to conducting systematic reviews, detailing the stages from study identification to inclusion and data synthesis.}
 \label{F9}
\end{figure} 

\textbf{Appendix B}

\begin{small}

{
\small
\setlength\LTleft{0pt}
\setlength\LTright{0pt}
\begin{longtable}{|
p{0.4cm}| 
p{4.2cm}| 
p{1.2cm}| 
p{1.2cm}| 
p{3.93cm}| 
p{3.93cm}|  
}
\caption{Recent Advances in Reinforcement Learning for Healthcare Applications: Categorization and Key Insights.}
\label{Table-Recent-Advances} \\
\hline
\rowcolor{headerpink}
\textbf{Ref.} &
\textbf{Key Concepts} &
\textbf{RL Algorithm} &
\textbf{Learning Type} &
\textbf{Outcome} &
\textbf{Challenges} \\
\hline
\endfirsthead
\hline
\rowcolor{headerpink}
\textbf{Ref.} &
\textbf{Key Concepts} &
\textbf{RL Algorithm} &
\textbf{Learning Type} &
\textbf{Outcome} &
\textbf{Challenges} \\
\hline
\endhead
	\cite{fatemi2022semi}& Applied semi-Markov Policy gradient methods to optimize warfarin dosing. & SDQN, SDDQN, SBCQ & Offline, MFRL & Optimized warfarin dosing by enhancing Time in Therapeutic Range (TTR) for stroke prevention. & Generalizability, computational complexity, overfitting, offline limitations. \\ 
		\hline
	\cite{tang2021model} & Framework for offline RL model selection in sepsis treatment. & FQI, FQE. & Offline, MFRL & Effective framework to select models for sepsis treatment & Simulated data, limited action space, ethical concerns. \\ 
		\hline
	\cite{yang2024deep}	 & Multi-class classification for COVID-19 screening and patient diagnosis in ICU. &DQN, D-DDQN & Offline, MFRL & Effective multi-class classification under imbalance. & Interpretability, trust issues, ethical and privacy concerns, computational complexity.\\ 
		\hline
	\cite{gutowski2023machine}	 & Optimizing individualized medicine intake schedules for Parkinson’s patients. & A3C (Actor Critic) & Online, Offline, MFRL & 7\% reduction in a cost function balancing motor symptom control and dosage regularity, & Simulated data limits generalizability, RL algorithm not explicitly stated. \\ 
		\hline
	\cite{tang2022leveraging}	 & Used factored action spaces in offline RL for improving sample efficiency and policy value. & Q-Learning & Offline, MFRL & Improved sample efficiency and policy value in decision-making. & Limited exploration in offline settings, lack of systematic study of factored action spaces. \\ 
		\hline
\cite{li2022electronic}	 & Cooperative multi-agent learning via linear value decomposition optimizing treatment strategies for diabetic ketoacidosis (DKA). & DQN. & Offline, MBRL & Improved BG control with a range of 70-150 mg/dl for DKA patients & Generalization, limited real-time applications, incomplete EHR data affects model’s generalization \\ 
		\hline
\cite{chen2022model}	 & Integrates discrete and continuous action handling for individualized ventilator control in ICU. & A3C & Offline, MBRL & Optimized ventilator settings with high data efficiency in ICU & Generalization and clinical adoption to real-time patient treatments \\ 
		\hline
\cite{job2024optimal}	 & Optimizing vasopressor policies for ARDS sepsis with DRL using LSTM-GRU state representation. & DDQN, DDDQN & Offline, MFRL & Optimized vasopressor policies for ARDS sepsis using DRL with LSTM-GRU & Limited patient data, synthetic data generation using Time GAN, computational complexity, and generalization concerns \\ 
		\hline
\cite{zhang2024optimizing}	 & Optimizing sepsis treatment with DDQN, reducing in-hospital mortality & DDQN & Offline, MFRL & Reduced in-hospital mortality by 3\% in sepsis treatment & Limited applicability to other databases, state heterogeneity, ethical privacy, and interpretability concerns. \\ 
		\hline
\cite{huang2022reinforcement} & Introduced continuous action RL for sepsis treatment. & DDPG and Twin Delayed DDPG (TD3) & Offline, MFRL & Optimized sepsis treatment, aligning with clinician decisions safely. & Extrapolation errors, overestimation in action performance, clinical safety, data generalizability, and evaluation challenges. \\ 
		\hline
\cite{wu2023value}	& Applies expert-informed value-based RL (WD3QNE) to personalize sepsis treatments. & WD3QNE & Offline, Rule based, MFRL & Achieved 97.81\% survival, outperforming other methods in sepsis treatment. & Generalization, clinical safety, and ethical concerns.\\ 
		\hline
\cite{wang2024deep}	 & Combining reinforcement learning with clinical and image data to support equitable and personalized treatment decisions. & DQN & Offline RL & Outperformed baseline clinical models in early lung cancer diagnosis rate (60.88\%) with lower error rates & Explainability and clinical scalability.\\ 
		\hline
\cite{jain2024irl} & IRL to optimize intervention scheduling in maternal and child health programs, using expert-derived preferences over restless bandit environments. & IRL & Offline RL, MFRL & Up to 260\% improvement in intervention allocation efficiency and 11\% increase in listener engagement, outperforming baselines. & Scaling IRL across thousands of agents, interpreting aggregate expert input, balancing fairness and performance, requiring iterative human feedback.\\ 
		\hline
\cite{yu2023deep} & Formulating cost-sensitive medical diagnosis as a multi-objective reinforcement learning problem to learn policies along the cost-accuracy Pareto frontier using clinical data. & SM-DDPO & Online, Offline, RL& Achieved state-of-the-art diagnostic accuracy with up to 80\% reduction in test cost, adaptable to new diseases without prior knowledge. & Non-decomposable objectives (e.g., F1 score), data imbalance, Pareto frontier computation, trade-off between interpretability and performance.\\ 
		\hline
\cite{choi2024deep}	 & Extracting optimal sepsis treatment policies using highlight-enhanced DDDQN trained on historical ICU records. & Highlight enhanced DDDQN & MoE (Offline \& Rule Based), MFRL & Up to 10.03\% and 9.81\% higher estimated survival rates on MIMIC-III and eICU datasets, vs actual treatments. Demonstrated generalizability across institutions. & Relies on estimated outcomes not verified in trials. Complex model behavior, data imputation, and safety concerns in offline RL settings.\\ 
		\hline
\cite{liu2024reinforcement}	 & Recommends personalized ventilator settings (PEEP, FiO2, tidal volume) using ICU data. Focused on safe offline learning from physician behavior. & BCQ & Offline, MFRL & Reduction in estimated mortality (eICU:12.1\%, MIMIC-IV:29.1\%) and improved oxygenation and BP vs. clinical baseline. Outperformed physician policies in simulation.& Offline-only evaluation; not prospectively validated. Relies on quality of historical data; real-world safety and interpretability need assessment.\\ 
		\hline
\cite{yang2024deep}	 & Imbalanced classification for COVID-19 prediction and ICU discharge prediction & DDDQN & Offline, MFRL & Improved sensitivity (0.806) with a 0.077 improvement for minority class prediction, especially for COVID-19 and ICU discharge tasks. & Lack of features for certain diagnoses (e.g., renal events), need for more data for multi-class tasks, and site-specific biases in datasets \\ 
		\hline
\cite{xu2024deep} & Semi-supervised segmentation of low-contrast and small objects in medical images using Deep generative Adversarial RL. & Soft Actor-Critic (SAC) & Offline, MFRL & Improved Dice: 85.02\% (brain tumor, +1.91\%), 73.18\% (liver tumor, +4.28\%), 70.85\% (pancreas, +2.73\%) & High computational cost, inefficient training, sparse reward issues.\\ 
		\hline
\cite{perera2025smart} & Learning from imperfect clinical decisions via a 2-phase framework using adversarial cooperative imitation learning and inverse RL. Applied to sepsis and diabetes. & ACIL, IRL, DQN & Offline, MFRL & Reduced sepsis mortality by 19.6\% compared to best baseline and HbA1c-High rates by 12.2\% in diabetes. Stronger alignment with clinical outcomes and policy effectiveness & Difficulty generalizing from imperfect data, complex reward design, and need for prospective clinical validation. \\ 
		\hline
\cite{nauka2025challenges} & Assessed transportability of the ICU-trained AI Clinician RL model to emergency department (ED) sepsis care; focused on 4 domains: missing data, measurement timing, diagnostic uncertainty, and treatment variability. & Policy Iteration (AI Clinician) & Offline, MFRL & $<1$\% had complete data by hour 6; 49\% received no treatment. & Delayed sepsis diagnosis, large treatment variation limit RL utility in ED. use of non deep RL may have led to poor feature abstraction, poor generalization and difficulty managing complex ED data.\\ 
		\hline
\cite{wang2024clinical} & Clinical knowledge-guided antibiotic combination recommendation for sepsis using RL. & DQN & Offline, MFRL & Achieved 79.09\% survival rate (vs. 60.2\% clinical baseline), 54.77\% of clinically deceased patients predicted to benefit from recommended treatments. Reduced average antibiotic duration from 9.48 to 5.52 days. & Generalization to other datasets, retrospective data limitations, missing data, limited features (e.g., lacking procalcitonin), need for personalization and multi-model decision systems.\\
\hline
\end{longtable}
}

\newpage 

   \begin{longtblr}[
		caption ={Clinical RL for Acute and Critical Care.},
		label = tab:Summary1-a 
		]{
			width = \linewidth,
			colspec = {|p{0.98cm}|p{1cm}|p{1.1cm}|p{0.8cm}|p{1.25cm}| p{4.6cm}|p{4.6cm}|},
			row{1} = {headerpink,c}, 
 cell{2}{1} = {r=3}{c},
 cell{2}{2} = {r=2}{c},
 cell{5}{1} = {r=10}{c},
 cell{5}{2} = {r=10}{c},
 cell{15}{1} = {r=3}{c},
 cell{15}{2} = {r=3}{c},
 vlines,
 hline{1-2,5,15,18} = {-}{},
 hline{3,6-14,16-17} = {3-7}{},
 hline{4} = {2-7}{},
		}
		Scenario & Sub-domain & Dataset & Sample Size & Algorithm & Research Outcome & Gaps \\
		\begin{sideways}\textbf{Mechanical Ventilation}\end{sideways} & \begin{sideways}ICU Weaning\end{sideways} & MIMIC-III,eICU & 11,943, 25,086 & FQI & 42.6\% reduced mortality, 83.3\% performance return & Data preprocessing, Limited generalizability across different geographic locations and Safety concerns.~\cite{peine2021development} \\
		& & MIMIC-III & 61532& CQL & Outperformed clinicians by 1.43\% in Ventilation Task & Overestimation of out-of-distribution actions, reward design challenges.~\cite{kondrup2023towards}\\ 
		& COVID Venti. & DDW & 3051 & DDQN& RL model improved ventilation safety with 56\% policy success in COVID-19 ICU patients & Sparse rewards, Dependence on domain expertise, overfitting risk and Limited generalizability~\cite{roggeveen2024reinforcement} \\ 
		\begin{sideways}\textbf{Sepsis Treatment}\end{sideways} & \begin{sideways}Dosage recommendation (Vaso and IV)\end{sideways} & MIMIC-III, AKI Dataset & 5783, 23950 & SM-DDPO & On Sepsis, F1=0.562, AUROC=0.845, cost reduced by 84\%. On AKI, F1=0.495, AUROC=0.795, cost reduced from $591 to $90. & Data Imbalance, Dependence on Label panels and Generalizability.~\cite{yu2023deep} \\
		& & MIMIC-III & 276, 232 & WD3QNE, DQN & WD3QNE (Human expertise) 97.81\% survival, D3QN 96.48\%, WD3QN 97.49\%, DQN 88.29\%, DDQN 88.33\%, return 23.63 &Issues with biases in data and model building, complexity of clinical knowledge integration and Limited Generalizability.~\cite{wu2023value}\\ 
		& & MIMIC-III & 19633 & DDPG & RL performed better Clinician-like decisions. & Offline RL extrapolation errors impact performance, Limited generalizability and Action space complexity.~\cite{huang2022reinforcement}\\ 
		& & MIMIC-III, eICU & 20927, 14875 & DDDQN & Vasopressor dosing showed 10.03\% higher survival in MIMIC-III, 9.81\% in eICU & Data extrapolation from historical data, and more research is needed on dosage recommendations, side effects, and ethical/privacy concerns in AI healthcare.~\cite{choi2024deep} \\
		& & MIMIC-IV & 71,272 & DQN & The model showed a 17.5\% improvement in survival rates for patients treated with the recommended regimen & Data bias, Lack of ground truth and ethical challenges in DRL healthcare trials.~\cite{zhang2024antibiotics}\\
		& & MIMIC-IV & 3535 & XGBoost & AUC 0.94, F1 score 0.937, improved sepsis mortality prediction with high accuracy. & Missing data, Feature limitations, Model complexity, Limited generalizability, and ethical and privacy issues.~\cite{zhang2024predicting} \\
		& & MIMIC-III & 7956 & DQN & RL model improved glucose control in DKA patients by 40\%, stabilizing levels in the optimal 80-140 mg/dL range. & Data generalization, clinical safety and real-time application remain an open challenge~\cite{li2022electronic} \\ 
		& & MIMIC-III & 2160 & DQN, DDQN & RL policy with vasopressor safety constraints reduced unsafe changes by 77.7\%, improving performance. Clinician policy: 15\% vasopressor use. Optimal: 38\%. Modified policy reduced dose changes, improving safety. Rewards: Clinician 7.16, Optimal 10.9, Modified 8.07 & Clinical safety, ethics, privacy, real-time application, and generalization are key concerns.~\cite{jia2020safe}\\
		& & MIMIC-IV, eICU-CRD & 5105, 21595 & BCQ & Hospital mortality rates reduced by 2.2\%, from 14.2\% to 12.1\%, based on estimated data & Retrospective study, ICU complexity, lack of validation, and low-resource generalizability. ~\cite{liu2024reinforcement}\\
		& & MIMIC-IV, eICU-CRD & 412 & RL-NN & RL-NN model recommended $\geq$ 30 mL/kg fluid resuscitation in 3 hours, reducing SOFA score by 23.71\%, improving outcomes & Retrospective study, ICU complexity, lack of validation, and low-resource generalizability. ~\cite{liang2025reinforcement} \\

    \begin{sideways}\textbf{Sedation Dosing}\end{sideways} & \begin{sideways}Sedative dosing \end{sideways} & MIMIC-III & 8,860 & FQI and BIRL & IRL captured clinician decisions, with 99.7\% consistency for ventilation and 54.2\% for sedative dosing, improving patient outcomes.& Preprocessing errors and reliance on a specific dataset limit generalizability. Real-time clinical implementation remains challenging due to regulatory and model complexity.~\cite{yu2019inverse} \\

     & & MIMIC-IV & 1,757 & DDPG &RL agent improved MAP by 26\%, sedation by 8\%, with reduced medication dosage, outperforming clinicians in sedation management.& Model relies on MIMIC-IV data, limiting generalizability. Clinical integration is complex, and long-term effects like drug habituation aren't considered.~\cite{eghbali2021patient} \\

     & & MIMIC-IV & 1,346 & DQN & RL agent outperformed clinicians, with 99.5\% RASS and 95.8\% MAP within target ranges, improving MPE by 28.55\% and 23.76\%, respectively.& Model's reliance on MIMIC-IV data limits generalizability. Delirium and long-term effects may not be fully addressed. Real-time implementation still needs validation.~\cite{eghbali2021patient} \\ 
	\end{longtblr}
\begin{longtblr}[
 caption = {Summary of key studies (Personalized Therapy)},
 label = tab:Summary3
]{
 width = \linewidth,
 colspec = {|p{0.98cm}|p{1.4cm}|p{1.2cm}|p{0.8cm}|p{1.25cm}| p{4.4cm}|p{4.2cm}|},
 row{1} = {headerpink, c},
 cell{2}{1} = {r=5}{c},
 vlines,
 hline{1-2,7} = {-}{},
 hline{3-6} = {2-7}{},
}
Senario & Sub-domain & Dataset & Sample Size & Algorithm & Research Outcome & Gaps\\ 
\begin{sideways}  \textbf{Radiotherapy} \end{sideways} &
Lung Cancer Care & Lung CT scans & 75 & SARSA & RL models improved IMRT planning for head and neck cancer: Left parotid $-6.96$Gy ($p<0.01$), Right parotid $-7.14 $Gy ($p < 0.01$), Oral cavity 0.10Gy ($p = 0.97$). Planning time: 13.58 minutes (min. 2.27 min, max. 44.82 min). & Training time (10,000+ iterations), limited control over planning variables, bias from small datasets, and lack of heterogeneity correction are key drawbacks.~\cite{stephens2024reinforcement} \\ 
&Stage III NSCLC Therapy & CT scans & 95 & D3QN & DRL optimised NSCLC radiation therapy, delivering 60 Gy in total in 42 days with personalised dose, outperforming clinical practice & The study relied on simulated data based on existing clinical protocols, which may not fully represent real-world variability and challenges in clinical practice.~\cite{tortora2021deep}\\
&Brachy therapy & ORNL-MIRD 1996 Phantom & N/A & MCNP & Brachytherapy for right lung cancer: Iridium-192 gave the highest dose (31.326 Gy), followed by Palladium-103 (6.925 Gy) and Caesium-131 (4.043 Gy). & Long simulation times with high N-Particle Source (NPS) values and the need for optimal seed numbers to minimize radiation exposure are potential concern of the research study.~\cite{purwati2025analysis}\\
&Prostate Cancer & MatRAD & N/A & DQN with CNN & The ``text-to-plan'' method achieved a mean reward of -211.88, outperforming RL (-259.26) with optimal dose distribution for prostate cancer. & The proposed Model struggles with 3D data understanding, relying on 2D slices and may generalise poorly in medical contexts without domain-specific adaptation. ~\cite{ferrante2025transforming}\\
&Lung Cancer Diag. & MIMIC-IV & 9982 & DQN & RL models improved IMRT planning, matching clinical plans for 44 Gy and 26 Gy doses while reducing organ doses and planning time & Lack of large-scale clinical trials, Challenges in optimising hyperparameters of the model.~\cite{wang2024clinical}
\end{longtblr}

\begin{longtblr}[
		caption = {Summary of key studies (Chronic Disease and Longitudinal Therapy)},
		label = tab:Summary2 
		]{
			width = \linewidth,
			colspec = {|p{1cm}|p{1.6cm}|p{1.7cm}|p{1cm}|p{1.25cm}| p{3.9cm}|p{3.9cm}|}, 
      row{1} = {headerpink,c}, 
  cell{2}{1} = {r=5}{c},
 cell{2}{2} = {c},
 cell{5}{2} = {c},
 cell{7}{1} = {c,b},
 cell{8}{1} = {r=7}{c,b},
 cell{10}{2} = {r=2}{c},
 cell{13}{2} = {r=2}{c},
 cell{15}{1} = {r=6}{c},
 cell{15}{2} = {c},
 vlines,
 hline{1-2,7-8,15,21} = {-}{},
 hline{3-6,9-10,12-13,16-20} = {2-7}{},
 hline{11,14} = {3-7}{},
		}
		Scenario & Sub-domain & Dataset & Sample Size & Algorithm & Research Outcome & Gaps\\
		\begin{sideways}\textbf{Diabetes Management}\end{sideways} & Insulin Dosing & EHR data & 20 & DQN & High-fat meals: 
		iAUC $38 \pm 223$, time below 3.9\% reduced to 1.8\%. Exercise meals: iAUC $132 \pm 181$, time below 3.9\% reduced to 1.4\%.  & Major challenges include a small sample size, lack of a control group, variability, short duration, unsupervised setting, and reliance on self-reported data.~\cite{jafar2024personalized}\\
		& Physical Activity & 126 RCT's & 6718 & Deep RL & 1,100 METs-min/week reduced HbA1c: Prediabetes 0.38\%-0.24\%, Controlled 0.47\%-0.40\%, Uncontrolled 0.64\%-0.49\%, Severe 1.02\%-0.66\%. & Challenges in small sample size, short intervention duration, exclusion of severe cases, and limited generalizability to non-Western populations and comorbidities. ~\cite{aguilera2024effectiveness}\\
		& Diabetes Control & MIMIC-IV, Diabetes dataset & 12305, 1601 & ACIL, IRL ad BC & SI-S2 reduced sepsis mortality by 19.6\%, and SI-D lowered HbA1c-High rates by 12.25\% compared to baselines & Limitations include sparse data in severe cases, reliance on imperfect clinician data, and challenges in generalising across different settings.~\cite{perera2025smart}\\
		& Diabetes Control & Simulated Dataset & 10 & D3QN & TIR increased: Scenario A 66.66\%-92.55\%, B 64.13\%-93.91\%, C 58.85\%-78.34\%. Glucose decreased; hypoglycemia reduced across scenarios. & Potential drawbacks include reliance on simulated data, limited generalizability, RL training constraints, and insulin sensitivity variability across patients. ~\cite{jaloli2024basal}\\
		& Artificial Pancreas Controller & Simulated Dataset & 30 & PPO & TIR 87.45\%, TAR 11.75\%, TBR 0.75\%, no severe hypo/hyperglycemia, cumulative reward $\geq 400$ after 4,000 episodes. & The study has limitations like low training efficiency, interpretability challenges, and reliance on simulation, with transferability concerns to real-world patients.~\cite{zhao2025safe}\\ 
 \begin{sideways} \textbf{ ~~~~~ HIV} \end{sideways} & Optimized Treatment Scheduling & Simulated data & N/A & DDQN & PER-DDQN outperformed other methods: 315 days for 5-day, 309 days for 1-day STI, with optimal drug schedules. & Drawbacks include using simulated data, high computational demands for one-day segments, limited interpretability, and potential issues with generalizing to other scenarios. ~\cite{yoon2024optimal}.\\ 
	\vspace{5cm} \begin{sideways}\textbf{Anticoagulant Dosing}\end{sideways} & Heparin in ICU & MIMIC-II& 4470 & DHMM & Improved aPTT, 42.6\% lower mortality, 83.3\% return & The model’s black-box nature, reliance on retrospective data, training inefficiency, and challenges in real-world clinical integration limit its practical use.~\cite{nemati2016optimal} \\
		& Warfarin for AF & RCT's & 28232 & BCQ with SMDP & RL dosing improved TTR by 6.78\%, reduced stroke/embolic events by 11\%, outperforming benchmark dosing. & Limited data and retrospective design, no real-time dosing adaptation. ~\cite{petch2024optimizing}\\
		& \vspace{1.5cm}\begin{sideways}Warfarin\end{sideways} & Simulated Data & 10000 & DQN & RL model outperforms baselines: PTTR for Normal 92.4\%, Sensitive 89.3\%, Highly Sensitive 90.5\%; higher doses for sensitive patients. & Reliance on virtual data, Small cohort sizes impact model performance. ~\cite{zadeh2023optimizing}\\
		& & NUH hospital data & 127 & CURATE. AI & CURATE.AI Quadratic model achieved 61\% ideal INR predictions, while Linear model reached 62.3\%, outperforming baselines. & Needs prospective validation, Inconsistent INR measurements, need for external validation, and challenges in real-world implementation.~\cite{gan2025ai}\\
		& AF Treatment & CAFR Study & 8540 & PDD-DQN & 64.98\% of clinician treatments matched RL model; model-concordant treatments reduced stroke, embolism, and death risk & Potential drawbacks include limited NOAC data, lack of full confounder adjustment, limited generalizability, and black-box nature of the RL model.~\cite{zuo2021improving}\\ 
		& \begin{sideways}Heparin\end{sideways} & MIMIC-III & 1911 & BCQ & Maintained therapeutic aPTT levels 87.7\% vs 55.6\% with traditional methods. & Missing data, Extrapolation errors in off-policy learning and lack of safety guarantees. ~\cite{lim2024development}\\
	\begin{sideways}\textbf{Anticoagulant Dosing}\end{sideways}	& & MIMIC-III & 500 & Latent BCL & RL model optimized heparin dosing, maintaining aPTT (60-100s), outperforming SAC/DDPG with improved dosing accuracy and stability.& Limited action space, Data quality, safety, Model complexity and Real-world adaptability.~\cite{qiu2022latent}\\
		\begin{sideways} \textbf{Drug Prescription} \end{sideways} & Chemo dosing & NLST dataset & 2500 & DQN & EarlyStop-RL achieved 60.88\% early diagnosis, 12.85\% false positives, 1.33\% false negatives, outperforming Lung-RADS and Brock & Reliance on simulated data, data imbalance in the test set, challenges in clinical integration, and limited interpretability of the model.~\cite{Wang2024}\\
		& Herbal medicine & Guang'anmen Hospital data & 5638 & DRQN & PrescDRL outperformed doctors: SSR, SCR, MCR up by 117\%-402\%, with improved precision (40.5\%), recall (63\%), and F1 score (51.5\%) & Challenges include longer treatment sequences, model interpretability issues, and the need for real-world clinical validation.~\cite{yang2023optimization}\\   
		& Immune system & Simulated Data & N/A & SIRL & Optimized immune system drug dosages, reducing drug costs and side effects & Reliance on simulated data, lack of noise handling, sensitivity to parameters, and model complexity are the major concerns.~\cite{chen2025event} \\
		& Substance use treatment & TEDS-Dataset & 3703076 & Contextual Bandits & Personalized treatment plans improved 1-year remission by 0.0735 (unconstrained), 0.0442 (cost-sensitive), vs. 0.0280 baseline. & Lack of longitudinal data, missing treatment duration, inaccurate relapse risk estimation, and limited granularity in outcome measurement are potential issues.~\cite{baucum2023optimizing} \\
		& Personalized Treatment & GDSC, CCLE & 1001, 1036 & PPO and A2C & PPORank outperformed others with NDCG of 0.7611 (CCLE) and ranked Lapatinib higher for HER2+, PARPi for mBRCA TNBC. & Drawbacks include reliance on cell line data, computational intensity for large datasets, limited generalizability for individual drugs, and lack of long-term patient data. ~\cite{liu2022deep} \\
		& Diabetes Drug & Simulated & 30 & PAE-POMDP & Effective-DQN outperformed DQN with better glucose control, higher rewards, and faster runtime (129 mins, 1513 MiB). & Reliance on simulated data, single-patient testing, hypoglycemia risk, action space limitations, and lack of real-world validation or drug synergy.~\cite{basu2023challenges}.
	\end{longtblr}

  \begin{longtblr}[
		caption = {Summary of key studies applying RL for clinical applications(Disease prediction and diagnosis,
    Surgical and robotic assistance pathological and rare diseases: A comparative overview}, 
		label=Tale-4_summary
		]{
			width = \linewidth,
			colspec = {|p{0.55cm}|p{1.5cm}|p{1.1cm}|p{1.4cm}|p{1.25cm}| p{4cm}|p{4.2cm}|},
			row{1} = {headerpink}, cell{1}{1} = {b},
 cell{2}{1} = {r=9}{c},
 cell{11}{1} = {r=2}{c},
 cell{13}{1} = {r=3}{c},
 cell{16}{1} = {r=6}{c},
 cell{22}{1} = {r=4}{c},
 cell{26}{1} = {r=2}{c},
 vlines,
 hline{1-2,11,13,16,22,26,28} = {-}{},
 hline{3-10,12,14-15,17-21,23-25,27} = {2-7}{},
		}
 Sce- nario  & SubDomain & \# Subjects & Data Source & RL Algorithm & Research Outcome & Gaps\\
		\begin{sideways}\textbf{Diagnostic Imaging and Screening}\end{sideways} & Coronary Artery Disease (CAD) & 41328 & EHR (APPROACH) registry & DQN and CQL & RL policies (QL, DQN, CQL) improved clinical outcomes by up to 32\%, with DQN and CQL ($\alpha=0.001$) showing the highest gains. & A promising RL approach to CAD treatment, but challenges remain with offline training, action space, and generalizability.~\cite{ghasemi2025personalized}\\
		& Cardiac Infarctions Prediction & 306 factors & Cardiac data & WTRL & The model achieved a probability of 86.04\% and conjecture of 85.82\%, with improvements in myocardial infarction factors. & Innovative RL approach shows promise, but simplifications may limit real-world applicability.~\cite{pradeep2025predicting}\\
		& Lung Cancer Detection & 2,500 & NLST trails & ES-RL & EarlyStop-RL outperformed models with 60.88\% early diagnosis, 12.85\% false positive, 1.33\% false negative, F1=0.85, MCC=0.80, NRI=0.24 (p<0.05). & EarlyStop-RL outperforms clinical models in early lung cancer diagnosis, but may face challenges with integration and computational complexity in real-world use.~\cite{wang2024deep}\\
		& Myocarditis Diagnosis & 18,869 & PTB-XL ECG & ELRL-MD & RLWBS outperforms CV-WBS and EE-WBS with 88.2\% F-measure, 90.6\% geometric mean, showing strong precision and adaptability in ECG diagnosis. & RLWBS excels in ECG classification with high accuracy, but depends on large labelled datasets and has computational demands for real-time application.~\cite{Xiao2025}\\
		& Arrhythmia Prediction & 47 & MIT-BIH & RL with CNN & RL model achieved 97.4\% accuracy in arrhythmia prediction. & Computational Complexity, Real-time adaptation and data variability are potential concerns of the study.~\cite{serhani2025enhancing}\\
		& Automated Disease Diagnosis & 30,000 & SymCat2 & HIRL & KI-CD improved diagnosis success by 7.1\%, disease classification by 8.04\%, SIR by 19.67\%, and AMR by 0.23\%. & KI-CD improves diagnosis, classification, and symptom identification but may face challenges with generalization and real-world complexity.~\cite{tiwari2022knowledge}\\
		& Respiratory failure Diagnosis & 12,519 & MIMIC-III & MHSAC & MHSAC outperformed with performance of 485.974, compared to 485.461 for HSAC and 346.801 for Discrete-SAC, lower standard deviation, and faster convergence (5 epochs). & MHSAC shows strong performance and stability, but may face challenges in real-world applications due to complexity and data reliance.~\cite{chen2022model}\\
		& Nonalcoholic Fatty Liver Disease (NAFLD) & N/A & NAFLD/ NASH & RL with DL & The proposed model improved NAFLD/NASH diagnosis with 94.2\% precision, 99.1\% AUROC in steatosis, and over 90\% AUROC in fibrosis and ballooning evaluation. & Need for further clinical validation, interobserver variability, and resource requirements for implementation.~\cite{takahashi2023artificial}\\
		& Alignment of LLMs with Radiologists & 94,844 & Chest CT & RLAIF with PPO & Proposed model showed 77.9\% agreement with radiologists, improving precision by 2.56\%, recall by 1.77\%, and F1 by 1.13\%. Metastasis precision: 87.1\%. & Data bias limits its applicability across diverse hospitals and languages remains an open challenge.~\cite{yang2025aligning}\\
		\begin{sideways}~~~~~~~~~~~~~~~~~~~~~~\textbf{Risk Scoring}\end{sideways} & Psychosis Prediction in Youth & 90,000 & Chest CT dataset & RLAIF With PPO and LLMs & Baichuan2 with LoRA achieved ROUGE-1: 0.6404, ROUGE-L: 0.5751, outperforming prompt-tuning in CT report summarization. & While LoRA provides a balance of performance and efficiency, full fine-tuning offers higher accuracy, but at a higher computational cost.~\cite{luther2025probabalistic}\\
		& Personalized Screening Policies & 43,749 & MGH & Envelope Q-learning & Tempo-Mirai achieved a 10\% increase in early detection and a 15\% reduction in screening frequency. & Tempo-Mirai improves early detection and reduces screening frequency, but its reliance on specific risk models may limit generalization to other populations.~\cite{yala2022optimizing}\\
    \begin{sideways}\textbf{Triage Systems}\end{sideways} & Triage System for Mass Casuality Incidents & 1204290 & NTDB & LR, RF, DNN & High triage accuracy (AUC 0.844), useful for MCI settings. & Good model performance but faces challenges with data imbalance, reliance on T-RTS, and the simplified Consciousness Scale affecting generalization~\cite{kim2021automated}.\\
		& Clinical Decision Support System & 248 & KHU data & CDSS & Sensitivity of 85\% for urgent cases; good usability (SUS score 78.2). & Poor sensitivity for non-urgent cases (19\%), missed diagnoses (41.8\%), and patient motivation issues, affecting generalizability.~\cite{maatta2023diagnostic}\\
		& Emergency Department Triage & 43 & Asclepius data & Llama-3-70b (MARL) & Multi-agent system accurately classified 85\% of KTAS levels, outperforming single-agent in clinical decision-making and critical findings. & Overestimated urgency, struggled with KTAS levels 3-5, and relied on incomplete data and web search for medical info are the critical concerns.~\cite{han2025development}\\
\begin{sideways}\textbf{Mental disorder and seizures diagnostics}\end{sideways} & Clinical High-Risk for Psychosis & 46 CHR, 51 healthy & G-PREP and NY-PREP & PRLT & CHR group showed impaired gain learning, particularly in 80\% probability ($p < 0.01$), linked to negative symptoms, psychosis risk, and lower functioning. &Small sample, cross-sectional design, and single task limit generalizability; depression could confound results.~\cite{luther2025probabalistic} \\ 
		& Alzheimer’s Disease with (Depression, Hypertension) & 1,736 & ADNI & Q-learning & RL-generated treatment regimens matched or outperformed clinicians at 100\% data utilization (Q-learning score: -2.82 vs. clinician’s policy score: -4.57) & Discrepancy in medication data, lack of neuroimaging, negative rewards, small sample size, and absence of active testing environment limit the study's accuracy.~\cite{bhattarai2023using} \\
		& Parkinson’s Disease & 1,000 & DBS (Neural data) & PPO & RL-based DBS reduced neuron synchronization more effectively than standard DBS; energy efficiency optimized (energy expended reduced from 14x standard DBS to 0.8x at 15 skipped steps & High computational demand, initial energy use 14x higher than DBS, reliance on simulations limits real-world applicability.~\cite{zhao2025reinforcement} \\
		& Alzheimer's Disease & 1,251 (419 AD, 832 CN) & ADNI & Q-Learning & RL model with Inception V3 improved AD classification (accuracy 90.2\%, AUC 0.95), sensitivity: 0.931, specificity: 0.908, outperforming previous methods by 5.1\%. & Class imbalance, high computational cost, and limited adaptability of data augmentation methods pose challenges for real-world applications.~\cite{hatami2024investigating} \\
		& Alzheimer's Disease & 160 & ADNI & PPO & The model predicted 10-year cognition trajectories, outperforming state-of-the-art methods with MAE 0.537 and MSE 0.761 on ADNI data. &The model used a simplified 2-node graph, had missing data issues, and relied on assumptions about brain activity, limiting its accuracy and generalizability.~\cite{saboo2021reinforcement} \\
		& Dementia Care & N/A & PLWD & DQN with LLMs & Proposed approach achieved average returns of 140, outperforming random (70), with verbal non-directive assistance at 144.3. & Simplified PLWD model, LLM-based behavior simulation prone to inaccuracies, and no real-world testing or evaluation by dementia care professionals.~\cite{yuan2025integrating} \\
		\begin{sideways}\textbf{Robotic Surgery and Assistance}\end{sideways} & Internal Tissue Points Manipulation & 20,000 & simulated & SA2C & The trained agent achieved 1.3mm error in tissue placement, with training time of 4.1hrs and planning time of 377s. & The method assumes flat tissue, may cause local damage, faces simulation-to-real gaps, and relies on detectable tissue points, limiting real-world applicability.~\cite{ou2023sim}\\
		& Surgical Automation (2D/3D) & N/A & simulated & SER & SER lowers constraint violations, improves task success convergence. 98.2\% real-world success (visual assist) & Computational cost, Sim-to-real transfer, Sparse rewards, Limited real-world safety validation ~\cite{fan2024safe}\\
		& Autonomous Blood Suction & 40 & simulated & DRL with LLM & LLM reasons blood suction order; accounts for bleeding, clots, tool proximity & LLMs struggle with real-time adaptation and real-world complexities.~\cite{zargarzadeh2025from} \\
		& Online trajectory guidance for novice surgeons & 10 & simulated & IL+AR & AR/IF reduces novice trajectory error by 76.47\%/65.15\%, narrowing expert gap in peg transfer task. & Generalisation, individual learning, computational cost, real-time, and haptic feedback integration remain open challenges.~\cite{chen2025online} \\
		\begin{sideways}\textbf{Pathological and Rare Diseases}\end{sideways} & Haemato- lymphoid
		diagnostics & N/A & EHR data & Q-Learning & RL model in lymphoma diagnostics improved accuracy, with AUROC 0.74 and better prognosis prediction. & Data scarcity for rare subtypes, integration challenges, and need for high-quality datasets are critical issues.~\cite{syrykh2025role} \\
		& Voice Pathology Detection & N/A & Synthetic & GRPO and PPO & GRPO with MoE achieved 98.60\% accuracy and 99.88\% AUC in voice pathology detection, but synthetic data limits real-world applicability. & The model showed high accuracy but relies on synthetic data, limiting generalizability. Real-world validation and model simplicity are needed.~\cite{togootogtokh2025voicegrpo} 
	\end{longtblr}
\end{small} 

\end{document}